\documentclass[10pt,journal,compsoc]{IEEEtran}

\ifCLASSOPTIONcompsoc
  \usepackage[nocompress]{cite}
\else
  \usepackage{cite}
\fi

\ifCLASSINFOpdf
\else
\fi

\usepackage{graphicx}

\usepackage{cite}
\usepackage{xcolor}
\usepackage[pagebackref=true,breaklinks=true,colorlinks,bookmarks=false]{hyperref}
\usepackage{amsmath}
\usepackage{amssymb}
\usepackage{multirow}
\usepackage{colortbl}
\usepackage{booktabs}
\usepackage{enumitem}
\usepackage{subfigure}
\usepackage{pdfpages}

\definecolor{mygray}{gray}{.85}

\DeclareRobustCommand{\rchi}{{\mathpalette\irchi\relax}}
\newcommand{\irchi}[2]{\raisebox{\depth}{$#1\chi$}} 

\newcommand{\tabincell}[2]{\begin{tabular}{@{}#1@{}}#2\end{tabular}}

\newcommand{\etal}{\textit{et al}.}
\newcommand{\ie}{\textit{i}.\textit{e}.}
\newcommand{\eg}{\textit{e}.\textit{g}.}
\newcommand{\vs}{\textit{vs}.}

\begin{document}
%

\title{Learning Signed Hyper Surfaces for Oriented Point Cloud Normal Estimation}

%
%
%

\author{
Qing Li~\IEEEmembership{Member, IEEE}, Huifang Feng, Kanle Shi, Yue Gao~\IEEEmembership{Senior Member, IEEE}, Yi Fang, Yu-Shen Liu~\IEEEmembership{Member, IEEE}, Zhizhong Han

\IEEEcompsocitemizethanks{
\IEEEcompsocthanksitem Qing Li, Yue Gao and Yu-Shen Liu are with the School of Software, Tsinghua University, Beijing, China. E-mail: \{gaoyue, liuyushen\}@tsinghua.edu.cn
\IEEEcompsocthanksitem Qing Li is with the School of Computing and Artificial Intelligence, Southwest Jiaotong University, Chengdu, China. E-mail: qingli@swjtu.edu.cn
\IEEEcompsocthanksitem Huifang Feng is with the School of Computer and Software Engineering, Xihua University, Chengdu, China. E-mail: fhf@xhu.edu.cn
\IEEEcompsocthanksitem Kanle Shi is with Kuaishou Technology, Beijing, China. E-mail: shikanle@kuaishou.com
\IEEEcompsocthanksitem Yi Fang is with the Center for Artificial Intelligence and Robotics, New York University Abu Dhabi, Abu Dhabi, UAE. E-mail: yfang@nyu.edu
\IEEEcompsocthanksitem Zhizhong Han is with the Department of Computer Science, Wayne State University, Detroit, USA. E-mail: h312h@wayne.edu
}

\thanks{The corresponding author is Yu-Shen Liu. This work was supported by the National Key R\&D Program of China (2022YFC3800600), the National Natural Science Foundation of China (62272263, 62072268), and in part by Tsinghua-Kuaishou Institute of Future Media Data.
The source code, data and pretrained models are available at \textcolor{red}{\href{https://github.com/LeoQLi/SHS-Net}{https://github.com/LeoQLi/SHS-Net}}.}
}

%
%

\markboth{Journal of \LaTeX\ Class Files,~Vol.~14, No.~8, August~2015}%
{Shell \MakeLowercase{\textit{et al.}}: Bare Advanced Demo of IEEEtran.cls for IEEE Computer Society Journals}
%



\IEEEtitleabstractindextext{
\begin{abstract}
  We propose a novel method called SHS-Net for point cloud normal estimation by learning signed hyper surfaces, which can accurately predict normals with global consistent orientation from various point clouds.
  Almost all existing methods estimate oriented normals through a two-stage pipeline, i.e., unoriented normal estimation and normal orientation, and each step is implemented by a separate algorithm.
  However, previous methods are sensitive to parameter settings, resulting in poor results from point clouds with noise, density variations and complex geometries.
  In this work, we introduce signed hyper surfaces (SHS), which are parameterized by multi-layer perceptron (MLP) layers, to learn to estimate oriented normals from point clouds in an end-to-end manner.
  The signed hyper surfaces are implicitly learned in a high-dimensional feature space where the local and global information is aggregated.
  Specifically, we introduce a patch encoding module and a shape encoding module to encode a 3D point cloud into a local latent code and a global latent code, respectively.
  Then, an attention-weighted normal prediction module is proposed as a decoder, which takes the local and global latent codes as input to predict oriented normals.
  Experimental results show that our algorithm outperforms the state-of-the-art methods in both unoriented and oriented normal estimation.
\end{abstract}

}

\maketitle

\IEEEdisplaynontitleabstractindextext

%
\IEEEpeerreviewmaketitle

\IEEEraisesectionheading{\section{Introduction}\label{sec:intro}}

\begin{figure}[t]
  \centering
  \includegraphics[width=\linewidth]{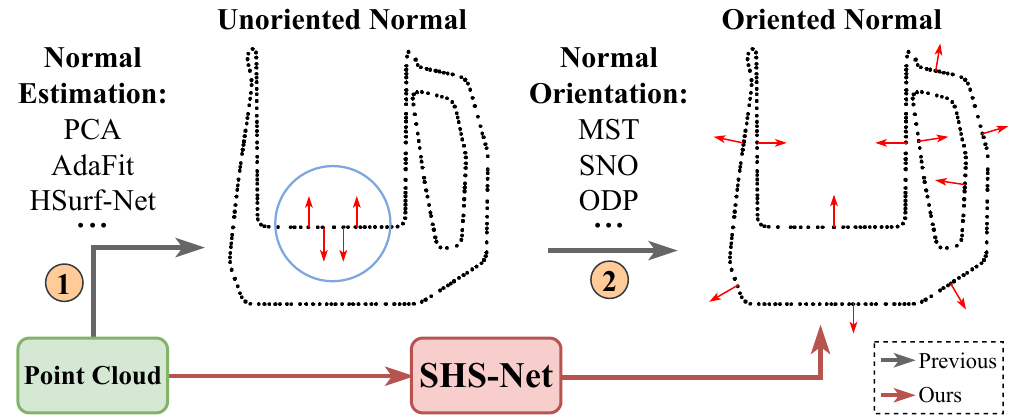}  \vspace{-0.6cm}
  \caption{
    We propose SHS-Net to estimate oriented normals directly from point clouds. In contrast, previous studies usually achieve this process through a two-stage paradigm using different algorithms, \ie, (1) unoriented normal estimation (\eg, PCA~\cite{hoppe1992surface}, AdaFit~\cite{zhu2021adafit} and HSurf-Net~\cite{li2022hsurf}) and (2) normal orientation (\eg, MST~\cite{hoppe1992surface}, SNO~\cite{schertler2017towards} and ODP~\cite{metzer2021orienting}).
  }
  \label{fig:intro}
  \vspace{-0.2cm}
\end{figure}

\IEEEPARstart{I}n computer vision and graphics, estimating normals for point clouds is a prerequisite for many techniques.
As an important geometric property of point clouds, normals with consistent orientation, \ie, \emph{oriented normals}, clearly reveal the geometric structures and make significant contributions in downstream applications, such as rendering and surface reconstruction~\cite{kazhdan2005reconstruction, kazhdan2006poisson, kazhdan2013screened}.
Generally, the estimation of oriented normals requires a two-stage paradigm (see Fig.~\ref{fig:intro}):
(1) the unoriented normal estimation from the local neighbors of the query point,
(2) the normal orientation to make the normal directions to be globally consistent, \eg, facing outward of the surface.
While unoriented normals can be estimated by plane or surface fitting of the local neighborhood, determining whether the normals are facing outward or inward is ambiguous.
In recent years, many excellent algorithms~\cite{lenssen2020deep,ben2020deepfit,zhu2021adafit,li2022graphfit,li2022hsurf} have been proposed for unoriented normal estimation, while there are few methods that have reliable performance for normal orientation or directly estimating oriented normals.
Estimating oriented normals from point clouds with noise, density variations, and complex geometries in an end-to-end manner is still a challenge.

The classic normal orientation methods rely on simple greedy propagation, which selects a seed point as the start and diffuses its normal orientation to the adjacent points via a minimum spanning tree (MST)~\cite{hoppe1992surface}.
These methods are limited by error accumulation, where an incorrect orientation may degenerate all subsequent steps during the iterative propagation.
Furthermore, they heavily rely on a smooth and clean assumption, which makes them easily fail in the presence of sharp edges or corners, density variations and noise.
Meanwhile, their accuracy is sensitive to the neighborhood size of propagation.
For example, a large size is usually used to smooth out outliers and noise, but can also erroneously include nearby surfaces.
Considering that local information is usually not sufficient to guarantee robust orientation, some improved methods~\cite{seversky2011harmonic,wang2012variational,schertler2017towards,xu2018towards,jakob2019parallel,metzer2021orienting} try to formulate the propagation process as a global energy optimization by introducing various constraints.
Since their constraints are mainly derived from local consistency, the defects are inevitably inherited, and they also suffer from cumulative errors.
Moreover, their data-specific parameters are difficult to generalize to new input types and topologies.

Different from the propagation-based methods, which only consider the adjacent normal orientation,
the volume-based approaches exploit volumetric representation, such as signed distance functions~\cite{mullen2010signing,mello2003estimating} and variational formulations~\cite{walder2005implicit,huang2019variational,alliez2007voronoi}.
They aim to divide the space into interior/exterior and determine whether point normals are facing inward or outward.
Despite improvements in accuracy and robustness, these methods cannot scale to large point clouds due to their computational complexity.
In general, propagation-based methods have difficulty with sharp features, while volume-based methods have difficulty with open surfaces.
Furthermore, the above-mentioned methods are usually complex and require a two-stage operation, their performance heavily depends on the parameter tuning in each separated stage.
Recently, several learning-based methods~\cite{guerrero2018pcpnet,hashimoto2019normal,wang2022deep} have been proposed to deliver oriented normals from point clouds and have exhibited promising performance.
Since they focus on learning an accurate local feature descriptor and do not fully explore the relationship between the surface's normal orientation and the underlying surface, their performance cannot be guaranteed across different noise levels and geometric structures.

In this work, we propose to estimate oriented normals from point clouds by implicitly learning \emph{signed hyper surfaces}, which are represented by MLP layers to interpret the geometric property in a high-dimensional feature space.
We learn this new geometry representation from both local and global shape properties to directly estimate normals with consistent orientation in an end-to-end manner.
The insight of our method is that determining a globally consistent normal orientation should require a global context to eliminate the orientation ambiguity in local regions since orientation should be related to the global structure.
We evaluate our method by conducting a series of qualitative and quantitative experiments on a range of point clouds with different sampling densities, noise levels, and thin and sharp structures.
We reported our original method in \cite{li2023shsnet} and extended our method with unoriented normal estimation, unoriented normal orientation, more applications, and experimental results.

Our main contributions can be summarized as follows.
\begin{itemize}[leftmargin=*]
\setlength{\itemsep}{3pt}
\setlength{\parsep}{0pt}
\setlength{\parskip}{0pt}
  \item We introduce a new technique to represent point cloud geometric properties as signed hyper surfaces in a high-dimensional feature space.
  \item We show that the signed hyper surfaces can be used to estimate normals with consistent orientations directly from point clouds, rather than through a two-stage paradigm.
  \item We also show that the modules we designed can be used to build a novel highly efficient pipeline with fewer parameters to estimate accurate unoriented normals, which can be combined with oriented normals to further improve our performance by using a new normal orientation strategy.
  \item We experimentally demonstrate that our method is able to estimate normals with high accuracy and achieves the state-of-the-art results in both unoriented and oriented normal estimation.
  \item We apply our method to downstream applications, such as surface reconstruction and point cloud filtering, and show that our estimated normals can effectively improve their performance. We also provide more analysis of the algorithm and experimental results on real-world indoor datasets based on the conference version.
\end{itemize}

\section{Related Work}

\subsection{Unoriented Normal Estimation}

\noindent\textbf{Traditional Methods}.
Over the past few decades, many algorithms have been proposed for point cloud normal estimation, such as the classic Principle Component Analysis (PCA)~\cite{hoppe1992surface} and its improvements~\cite{alexa2001point,pauly2002efficient,mitra2003estimating,lange2005anisotropic,huang2009consolidation}.
Generally, according to Singular Value Decomposition (SVD)~\cite{stewart1993early}, the covariance matrix of a local patch is decomposed and the eigenvector with the smallest eigenvalue is perpendicular to the plane defined by the patch.
Thanks to its simplicity and efficiency, PCA is widely used in various point cloud processing tasks.
However, it is always difficult to determine the data-specific parameter, \eg, patch size, which is crucial to the accuracy of estimation.
To find an optimal size for different data, Mitra \etal~\cite{mitra2003estimating} propose to costly investigate the effect of local curvature and point density of the underlying surface.
Later, some works introduce Hough transform~\cite{boulch2012fast} and Voronoi-based paradigms~\cite{amenta1999surface,merigot2010voronoi,dey2006provable,alliez2007voronoi} to improve the robustness of normal estimation and deal with sharp features.
Furthermore, the pattern description of local patches is not limited to planes, various complex surfaces~\cite{levin1998approximation,cazals2005estimating,guennebaud2007algebraic,aroudj2017visibility,oztireli2009feature} are adopted to more accurately fit the surface represented by the point cloud, such as moving least squares~\cite{levin1998approximation}, truncated Taylor expansion ($n$-jet fitting)~\cite{cazals2005estimating} and spherical surface fitting~\cite{guennebaud2007algebraic}.
These traditional methods are usually sensitive to noise and various data types, and have limited accuracy even with heavy fine-tuned parameters.

\noindent\textbf{Learning-based Methods}.
More recently, learning-based methods have been proposed to improve performance in this area and can be mainly divided into two categories: regression-based and surface fitting-based.

(1) \emph{Regression-based methods}.
The regression-based methods try to directly predict normals from structured data~\cite{boulch2016deep,roveri2018pointpronets,lu2020deep} or raw point clouds~\cite{guerrero2018pcpnet,zhou2020normal,hashimoto2019normal,ben2019nesti,zhou2020geometry,zhou2022refine,li2022hsurf,li2023NeAF} in a data-driven manner.
For example, HoughCNN~\cite{boulch2016deep} uses the Hough transform to convert 3D points into 2D grid representations, and then trains a simple neural network to select a normal from a Hough image-accumulator.
PCPNet~\cite{guerrero2018pcpnet} is regarded as the prior work that adopts the PointNet architecture~\cite{qi2017pointnet} to extract patch features and predict point normals and curvatures.
Based on PCPNet, Zhou \etal~\cite{zhou2020normal} introduce a local plane constraint and a multi-scale neighborhood selection strategy.
Nesti-Net~\cite{ben2019nesti} aims to learn a multi-scale feature vector, and tries to costly find the optimal neighborhood scale for each point.
HSurf-Net~\cite{li2022hsurf} achieves good performance by learning hyper surfaces from local patches, but the learned surfaces have no sign and cannot determine the normal orientation.
NeAF~\cite{li2023NeAF} selects normals from randomly sampled vectors by predicting the angular offset of the query vector.
MSECNet~\cite{xiu2023msecnet} improves normal estimation in areas with drastic normal changes by introducing edge detection technology.
CMG-Net~\cite{wu2024cmg} proposes a metric of Chamfer Normal Distance to address the issue of normal direction inconsistency in noisy point clouds.

(2) \emph{Surface fitting-based methods}.
The surface fitting-based methods integrate the traditional surface fitting techniques, such as plane fitting~\cite{lenssen2020deep,cao2021latent} and jet fitting~\cite{ben2020deepfit,zhu2021adafit,zhou2023improvement,zhang2022geometry,li2022graphfit,du2023rethinking}, into the end of the learning pipeline.
They usually carefully design a network to predict pointwise weights, and then use a weighted surface formulation to solve the normal of the fitted surface.
For example, Lenssen \etal~\cite{lenssen2020deep} propose to iteratively refine a weighted least squares plane fitting by introducing an adaptive anisotropic kernel.
MTRNet~\cite{cao2021latent} aims to fit a latent tangent plane by designing a differentiable RANSAC-like module.
DeepFit~\cite{ben2020deepfit}, AdaFit~\cite{zhu2021adafit}, GraphFit~\cite{li2022graphfit}, Zhang \etal~\cite{zhang2022geometry}, Zhou \etal~\cite{zhou2023improvement} and Du \etal~\cite{du2023rethinking} predict pointwise weights of local neighborhoods through a PointNet or graph convolutional network, and then apply a weighted polynomial surface fitting to calculate the surface normal.
The unoriented normals estimated by the above-mentioned methods randomly face both sides of the surface and cannot be used in many downstream applications without normal orientation.

\subsection{Consistent Normal Orientation}

To make the unoriented normals have globally consistent orientations, early approaches mainly focus on local consistency and use the orientation propagation strategy upon a minimum spanning tree (MST) to let the adjacent points have the same orientations, such as the pioneering work of~\cite{hoppe1992surface} and its improved methods~\cite{konig2009consistent,seversky2011harmonic,wang2012variational,schertler2017towards,xu2018towards,jakob2019parallel}.
These methods have many limitations in real applications as we introduced earlier.
Typical examples are that noisy and sharp features may lead to incorrect orientation propagation, and local errors will spread to larger regions and eventually result in severe performance degradation.
In addition, the orientation step cannot correct the wrong directions of the initial unoriented normal (\eg, orthogonal to the true normal).
Wang \etal\cite{wang2012variational} present a variational model that makes normals perpendicular to and consistent along the shape surface by minimizing a combination of the Dirichlet energy and the coupled-orthogonality deviation.
Their method requires parameter fine-tuning for complex features and may fail on data with outliers.
Schertler \etal~\cite{schertler2017towards} formulate the orientation process as a graph-based energy minimization problem, which is solved by improved quadratic pseudo-Boolean optimization~\cite{rother2007optimizing}.
However, the accuracy of their orientation depends heavily on the chosen normal flipping criterion.
Jakob \etal~\cite{jakob2019parallel} perform a graph-based energy optimization on the GPU for the entire point cloud.
It uses a parallel greedy solver to achieve faster speed than previous works.
Although the above works propose different improved flip criteria or formulate orientation as various global optimization problems to reduce the failure rate of orientation inversion, it is still difficult to guarantee robustness to different inputs.
ODP~\cite{metzer2021orienting} aims to achieve global consistency by introducing a dipole propagation strategy across the partitioned patches, but its robustness may suffer from the patch partition of nearby gaps or nested structures.
GCNO~\cite{xu2023globally} proposes to characterize the requirements of an acceptable winding-number field.
The oriented normals are found by utilizing these requirements to ensure global consistency and relying on the Voronoi diagram to estimate normals.
However, its optimization is extremely time-consuming for a large number of points since it needs to repeatedly evaluate the winding number of each data point and each query point.
NGLO~\cite{li2023NGLO} first predicts coarse normals with global consistency from the whole point cloud by learning implicit functions, and then refines the normals based on local information to improve their accuracy.
In contrast, some other approaches~\cite{guerrero2018pcpnet,hashimoto2019normal,wang2022deep} explore to gather information of different scales and directly predict oriented normals through end-to-end deep networks.
These methods focus on learning a general mapping from point clouds to normals and neglect the underlying surface distribution for normal orientation, leading to a sub-optimal solution.
In conclusion, the global orientation of point cloud normals is still an open problem with much room for improvement.

The task of implicit unoriented reconstruction, \ie, reconstructing surfaces from point clouds without normals, is also closely related to normal orientation, where the orientation and the reconstruction are bridged in implicit space~\cite{xiao2023point}.
Specifically, some approaches propose to solve the consistent normal orientation through volumetric representation.
They are usually developed for reconstructing surfaces from unoriented points by various techniques, such as signed distance functions~\cite{mullen2010signing,mello2003estimating}, variational formulations~\cite{walder2005implicit,huang2019variational,alliez2007voronoi}, visibility~\cite{katz2007direct,chen2010binary} and active contours~\cite{xie2004surface}.
Xiao \etal~\cite{xiao2023point} propose to incorporate isovalue constraints to the Poisson equation, and optimize implicit functions and point normals simultaneously.
iPSR~\cite{hou2022iterative} runs Poisson reconstruction in an iterative manner and updates normals using the generated surface of the last iteration.
PGR~\cite{lin2022surface} takes the point normal and surface element in the Gauss formula as unknown parameters, and then optimizes the parametric function space.
In addition to the traditional methods mentioned above, some other works~\cite{atzmon2020sal,gropp2020implicit,ma2020neural,Zhou2022CAP-UDF,ma2022reconstructing} use deep neural networks to learn implicit surfaces directly from raw point clouds without using training labels.
We know that the gradient determines the direction of function convergence, and the gradient of the iso-surface can be used as the normal of the surface.
Some methods add normals to constraints during optimization to assist surface reconstruction.
For example, SAP~\cite{peng2021shape} proposes a differentiable Poisson solver to represent shape surfaces as oriented point clouds, and the point positions and normals are updated during the optimization of surface.
Neural-Pull~\cite{ma2020neural} predicts the signed distance field to move a point along or against the gradient for finding its nearest path to the surface, and its gradient is equivalent to normal.
IGR~\cite{gropp2020implicit} proposes an implicit geometric regularization to encourage unit norm gradients and favor a smooth zero-level set of an implicit function.
Experimental results show that these methods have limited ability to deal with noise.
If the gradient is properly guided, the convergence can be robust and efficient, avoiding local extremum caused by noise or outliers.
To handle different types of data, effective gradient constraints need to be further explored.

\begin{figure*}[t]
  \centering
  \includegraphics[width=0.95\linewidth]{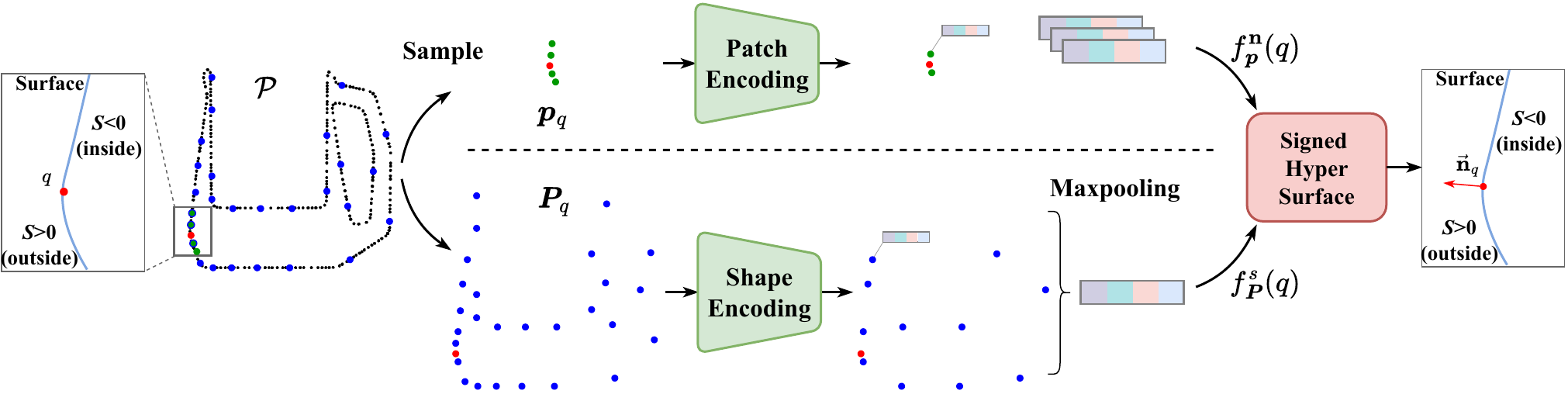}  \vspace{-0.25cm}
  \caption{
    The learning pipeline of the signed hyper surfaces for oriented normal estimation.
    It consists of two parallel branches, \ie, patch encoding and shape encoding, which have similar network architectures (see Fig.~\ref{fig:feat}), to extract local and global latent codes, respectively.
    In both branches, the number of point clouds is downsampled relative to the query point $q$.
    Finally, an embedding of the signed hyper surface is used to regress the oriented normal of the query point, which points to the outside of the shape surface.
  }
  \label{fig:net}
  \vspace{-0.2cm}
\end{figure*}

\begin{figure}[t]
  \centering
  \includegraphics[width=\linewidth]{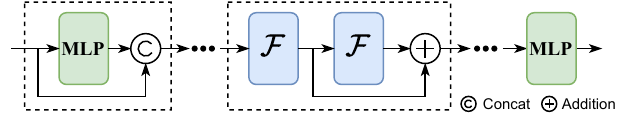}  \vspace{-0.5cm}
  \caption{
    Feature encoding network in patch and shape encoding.
    $\mathcal{F}$ is the latent code extraction layer.
    Black dots indicate the repetition of blocks (dashed box).
  }
  \label{fig:feat}
  \vspace{-0.2cm}
\end{figure}

\section{Preliminary}  \label{sec:pre}

In mathematics, an explicit representation in Euclidean space expresses the $z$ coordinate of a point $p$ in terms of the $x$ and $y$, \ie, $z \!=\! f (x, y)$.
Such a surface is called an explicit surface, also called a height field.
Another symmetric representation is $F(x, y, z) \!=\! 0$, where $F$ implicitly defines a locus called an implicit surface, also called a scalar field~\cite{bloomenthal1997introduction}.
The implicit surface is a zero iso-surface of $F$, \ie, the point set $\{p \in \mathbb{R}^3 : F(p) \!=\! 0\}$ is a surface implicitly defined by $F$.
Sampling points from an implicit surface is difficult, but the relationship between points and surfaces can be easily determined.
On the contrary, it is easy to sample points from an explicit surface, but it is difficult to determine the relationship between the points and the surface.
The explicit surface is usually used in surface fitting-based normal estimation, such as jet fitting~\cite{cazals2005estimating}, while the implicit surface is widely used in surface reconstruction.
Generally, an explicit surface, \ie, $z \!=\! f(x,y)$, can always be rewritten as an implicit surface, \ie, $F(x,y,z) \!=\! z-f(x,y) \!=\! 0$.
These two surface representations have the same tangent plane at a given point, where the normal is defined.

\noindent
\textbf{Explicit Surface Fitting}.
We employ the widely used $n$-jet surface model~\cite{cazals2005estimating} to briefly review the explicit surface fitting for normal estimation.
It represents the surface by a polynomial function $J_{n}:\mathbb{R}^2 \!\to\! \mathbb{R}$, which maps a coordinate $(x,y)$ to its height $z$ that is not in the tangent space by
\begin{equation}
  z \doteq J_{\alpha,n}(x,y) = \sum_{k=0}^{n} \sum_{j=0}^{k} {\alpha}_{k-j, j} x^{k-j} y^{j},
\end{equation}
where $\alpha$ is the coefficient vector that defines the surface function.
In order to find the optimal solution, the least squares approximation strategy is usually adopted to minimize the sum of the square errors between the (ground truth) height and the jet value over a point set $\{ p_i\}_{i=1}^N$,
\begin{equation} \label{eq:ls}
  J_{\alpha,n}^{\ast} = \mathop{\rm argmin}_{\alpha} \sum_{i=1}^{N} \|z_i - J_{\alpha,n}(x_i,y_i) \|^2.
\end{equation}
If $\alpha \!=\! (\alpha_{0,0},\alpha_{1,0},\alpha_{0,1},\cdots,\alpha_{0,n})$ is solved, then the normal at point $p$ on the fitted surface is computed by
\begin{equation} \label{eq:fit_normal}
  \mathbf{n}_{p} = h(\alpha) = (-\alpha_{1,0}, -\alpha_{0,1}, 1) / \sqrt{1 + \alpha_{1,0}^2 + \alpha_{0,1}^2} ~~.
\end{equation}

\noindent
\textbf{Implicit Surface Learning}.
In recent years, many learning-based approaches have been proposed to represent surfaces by implicit functions, such as signed distance function (SDF)~\cite{park2019deepsdf} and occupancy function~\cite{mescheder2019occupancy}.
The signed (or oriented) distance function is the shortest distance of a given point $p\!=\!(x_{0},y_{0},z_{0})$ to the closest surface ${\boldsymbol S}$ in a metric space, with the sign determined by whether the point is inside ($F(p)\!<\!0$) or outside ($F(p)\!>\!0$) of the surface.
The underlying surface is implicitly represented by the iso-surface of $F(p)\!=\!0$.
In the surface reconstruction task, a deep network is usually adopted to encode a 3D shape into a latent code, which is fed into a decoder together with query points to predict signed distances.
If an implicit surface function is continuous and differentiable, the formula of tangent plane at a regular point $p$ (gradient is non-null) is $F_x(p)(x-x_0)+F_y(p)(y-y_0)+F_z(p)(z-z_0) \!=\! 0$ and its normal (\ie, perpendicular) is $\mathbf{n}_p\!=\!\nabla F(p) / \|\nabla F(p)\|$.

\section{Method}

As shown in Fig.~\ref{fig:net}, we propose to implicitly learn signed hyper surfaces in the feature space for estimating oriented normals.
In the following sections, we first introduce the representation of signed hyper surfaces by combining the characteristics of the above two surface representations.
Then, we design an attention-weighted normal prediction module to solve the oriented normals of query points from signed hyper surfaces.
Finally, we introduce how to learn this new surface representation from patch encoding and shape encoding using our designed loss functions.

\subsection{Signed Hyper Surface}  \label{sec:sgn_surf}

Similar to the learning of implicit surface, the signed hyper surface is implicitly learned by taking the latent encodings of point clouds as inputs and outputting an approximation of the surface in feature space,
\begin{equation}
    f_{\boldsymbol S}(\rchi) \approx \mathcal{E}_{\theta}(\rchi | z_1, z_2),~ z_1 = e_{\varphi}({\boldsymbol P}^1_{\rchi}),~ z_2 = e_{\psi}({\boldsymbol P}^2_{\rchi}),
\end{equation}
where $\mathcal{E}$ is implemented by a neural network with parameter $\theta$ that is conditioned on two latent vectors $z_1,z_2 \!\in\! \mathbb{R}^c$, which are extracted from point clouds by encoders $e_{\varphi}$ and $e_{\psi}$, respectively.
${\boldsymbol P}^1_{\rchi}$ and ${\boldsymbol P}^2_{\rchi}$ are subsample sets of the raw point cloud $\mathcal{P}$, \eg, point patches around a given point $\rchi$.

Similar to existing unoriented normal estimation methods~\cite{li2022hsurf,ben2020deepfit,zhu2021adafit,guerrero2018pcpnet},
we use a local patch ${\boldsymbol p}_q$ to capture the local geometry for accurately describing the surface pattern around a query point $q$,
\begin{equation}
    f^{\mathbf{n}}_{\boldsymbol p}(q) = \mathcal{E}^{\mathbf{n}}_{\theta}(q | z^{\mathbf{n}}_q),~
    z^{\mathbf{n}}_q = e_{\varphi}({\boldsymbol p}_q).
\end{equation}
Since the interior/exterior of a surface cannot be determined reliably from a local patch, we take a global subsample set ${\boldsymbol P}_q$ from the point cloud $\mathcal{P}$ to provide additional information to estimate the sign at point $q$,
\begin{equation}
    f^s_{\boldsymbol P}(q) = \mathrm{sgn}\big(g^s(q)\big) = \mathrm{sgn}\big(\mathcal{E}^{s}_{\theta}(q | z^s_q)\big),~
    z^s_q = e_{\psi}({\boldsymbol P}_q),
\end{equation}
where $\mathrm{sgn}(\cdot)$ is signum function, $g^s(q)$ denotes logit of the probability that $q$ has a positive sign.
Thus, the signed hyper surface function at point $q$ is formulated as
\begin{equation}
  f_{\boldsymbol S}(q) = f^{\mathbf{n}}_{\boldsymbol p}(q) \cdot f^s_{\boldsymbol P}(q) = \mathcal{E}^{\mathbf{n},s}_{\theta}(q | z^{\mathbf{n}}_q, z^s_q).
\end{equation}
Different from the surface reconstruction task that learns SDF by representing a surface as the zero-set of the SDF, we do not learn a distance field of points with respect to the underlying surface.

\subsection{Oriented Normal Estimation}  \label{sec:nor_esti}

To simplify notations, we denote $\mathcal{E}^{\mathbf{n},s}_{\theta}(q | z^{\mathbf{n}}_q, z^s_q)$ as $\mathcal{S}_{\theta}(\mathcal{X},\mathcal{Y})$, where $z^{\mathbf{n}}_q \!=\! \mathcal{X} \!\in\! \mathbb{R}^c$ and $z^s_q \!=\! \mathcal{Y} \!\in\! \mathbb{R}^c$ are high dimensional latent vectors.
According to the explicit surface fitting,
we formulate the signed hyper surface $\mathcal{S}_{\theta}:\mathbb{R}^{2c} \!\to\! \mathbb{R}^c$ as a feature-based polynomial function~\cite{li2022hsurf}
\begin{equation}  \label{eq:poly}
  \mathcal{S}_{\theta,\mu}(\mathcal{X},\mathcal{Y}) = \sum_{k=0}^{\mu} \sum_{j=0}^{k} \theta_{k-j,j} ~ \mathbf{x}_{k-j} \mathbf{y}_j = \theta ~ [\mathcal{X} : \mathcal{Y}],
\end{equation}
where [~:~] means the feature fusion through concatenation, $\mu$ denotes the number of fused items.

Similar to Eq.~\eqref{eq:ls}, the bivariate function $\mathcal{S}_{\theta,\mu}(\mathcal{X},\mathcal{Y})$ aims to map a feature pair $(\mathcal{X}_i, \mathcal{Y}_i)$ to their ground truth value $\mathcal{Z}_i=\hat{\mathcal{S}}(\mathcal{X}_i, \mathcal{Y}_i) \in \mathbb{R}^c$ in the feature space, \ie,
\begin{equation}
  \mathcal{S}_{\theta,\mu}^{\ast} = \mathop{\rm argmin}_{\theta,\mu} \sum_{i=1}^{N} \| \mathcal{Z}_i - \mathcal{S}_{\theta,\mu}(\mathcal{X}_i,\mathcal{Y}_i) \|^2 .
\end{equation}
To solve the oriented normal $\vec{\mathbf{n}}$ from signed hyper surfaces, we introduce a normal prediction module $\mathcal{H}(\cdot)$, thus
\begin{equation} \label{eq:normal_out}
  \mathcal{S}_{\theta,\mu}^{\ast} = \mathop{\rm argmin}_{\theta,\mu} \sum_{i=1}^{N} \| \mathcal{H}(\mathcal{Z}_i) - \mathcal{H}(\mathcal{S}_{\theta,\mu}(\mathcal{X}_i,\mathcal{Y}_i)) \|^2 .
\end{equation}
Finally, the oriented normal is optimized by
\begin{equation} \label{eq:loss}
  \mathcal{S}_{\theta,\mu}^{\ast} = \mathop{\rm argmin}_{\theta,\mu} \sum_{i=1}^{N} \| \hat{\vec{\mathbf{n}}}_i - \vec{\mathbf{n}}_i \|^2.
\end{equation}

\begin{figure}[t]
  \centering
  \includegraphics[width=\linewidth]{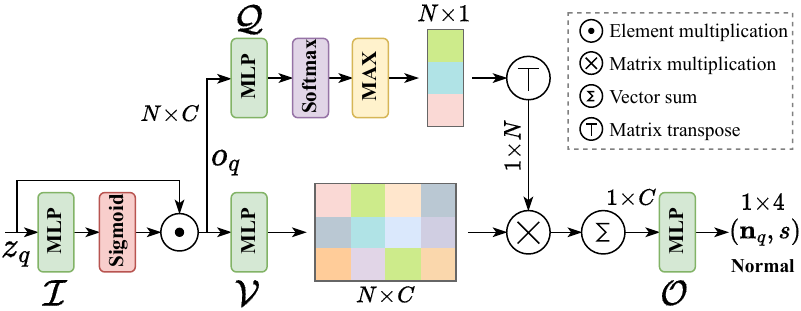}  \vspace{-0.65cm}
  \caption{
    Attention-weighted normal prediction module $\mathcal{H}(\cdot)$.
    After we obtain the surface embedding $z_{q}$ from the fused local and global latent code $[z^{\mathbf{n}}_q : z^s_q]$, we can predict the normal $\mathbf{n}_q$ of the query point $q$ and the sign $s$ to determine its orientation.
  }
  \label{fig:atten}
  \vspace{-0.2cm}
\end{figure}

\noindent
\textbf{Attention-weighted Normal Prediction $\mathcal{H}(\cdot) \!:\! \mathbb{R}^c \!\to\! \mathbb{R}^4$}.
As shown in Fig.~\ref{fig:atten}, we use an attention mechanism to recover the oriented normal $\vec{\mathbf{n}}_q$ of the query point $q$ from $c$-dimensional fused surface embedding $z_{q}$,
\begin{equation}  \label{eq:output}
  (\dot{\mathbf{n}}_q, s) \!=\! \mathcal{O}\big( \mathcal{V}(o_{q}) \otimes {\rm MAX} \big\{ {\rm softmax}_{\mathcal{N}_q} \big( \mathcal{Q}_j(o_{q})_{j=1}^m \big) \big\} \big),
\end{equation}
where $o_{q} \!=\! \tau \cdot z_{q}, \tau \!=\! {\rm sigmoid}(\mathcal{I}(z_{q}))$. $\mathcal{O}, \mathcal{V}, \mathcal{Q}$ and $\mathcal{I}$ are MLPs.
$m\!=\!64$ is the feature dimension size.
First, a multi-head strategy is adopted to deliver $m$ relative weights $\mathcal{Q}_j(o_{q})$, which are normalized by softmax over neighbors $\mathcal{N}_q$ into positive interpolation weights.
Then, the feature maxpooling ${\rm MAX}\{\cdot\}$ is performed to produce attention weights for each point.
Meanwhile, the feature embedding $o_q$ is refined through another branch $\mathcal{V}$ and modulated as the weighted sum through matrix multiplication.
Finally, the normal and its sign (\ie, orientation) $\vec{\mathbf{n}}_q \!=\! (\mathbf{n}_q \!\in\! \mathbb{R}^3, s \!\in\! \mathbb{R})$ is predicted as a 4D vector by $\mathcal{O}$, and $\mathbf{n}_q \!=\! {\dot{\mathbf{n}}_q} / {\|\dot{\mathbf{n}}_q\|}$.

\subsection{Feature Encoding}

\noindent
\textbf{Patch Encoding}.
Given a neighborhood point patch ${\boldsymbol p}_q$ of the query point $q$, our local latent code extraction layer $\mathcal{F}$ is formulated as
\begin{equation} \label{eq:local}
    \dot{z}^{\mathbf{n}}_{i} \!=\! \mathcal{A} \left(\mathcal{B} \left({\rm MAX} \big\{ \mathcal{C} (w_{j} \cdot z^{\mathbf{n}}_{j}) \big\}_{j=1}^{N_{l}} \right), z^{\mathbf{n}}_{i} \right),
\end{equation}
where $i\!=\!1,\cdots,N_{l+1}$, $l$ is the neighborhood scale index and $N_{l+1} \!\leqslant\! N_{l}$.
$z^{\mathbf{n}}_{i}\!=\!\mathcal{D}(p_i), p_i \!\in\! {\boldsymbol p}_q$ is the per-point feature in the patch.
$\mathcal{A}, \mathcal{B}, \mathcal{C}$ and $\mathcal{D}$ are MLPs.
${\rm MAX}\{\cdot\}$ denotes the feature maxpooling over $N_{l}$-nearest neighbors of the query point $q$.
$w$ is a distance-based weight given by
\begin{equation}  \label{eq:weight}
    w_{j} = \frac{\beta_j}{\sum_{i=1}^{N}{\beta_i}}, ~ \beta_i = {\rm sigmoid} \big(\gamma_1 - \gamma_2 {||p_i - q||}_2 \big),
\end{equation}
where $\gamma_1$ and $\gamma_2$ are learnable parameters with an initial value of $1.0$.
We use the weight $w$ to make the layer focus on the points $p_i$ that are closer to the query point $q$ in areas where the geometry changes drastically, thereby improving the robustness of feature encoding.
As shown in Fig.~\ref{fig:feat}, we stack two layers $\mathcal{F}$ to form a block, which is further stacked to build our patch feature encoder $e_{\varphi}$.

\noindent
\textbf{Shape Encoding}.
Since the global subsample set ${\boldsymbol P}_q \!=\! \{p_i\}_{i=1}^{N_{\boldsymbol P}}$ can be seen as a patch with points distributed globally on the shape surface,
we adopt a similar network architecture with the patch feature encoder to get the global latent code $z^s_q$.
To obtain ${\boldsymbol P}_q$, we use a probability-based sampling strategy~\cite{erler2020points2surf}, which brings more points closer to the query point $q$.
It samples points according to a density gradient that decreases with increasing distance from the point $q$.
Moreover, we find that adding some points from uniform sampling can bring better results in structures with different densities and concavities.
Then, the gradient of a point is calculated by
\begin{equation}  \label{eq:sample}
  \upsilon (p_i) = \left\{
  \begin{aligned}
  & \left[ 1 - 1.5 \frac{\|p_i-q\|_2}{\max_{p_j \in \mathcal{P}} \|p_j-q\|_2} \right]_{0.05}^1 ~;\\
  & ~1 ~~~\text{if}\ i \in \mathcal{R}~,
  \end{aligned}
  \right.
\end{equation}
where $[\cdot]_{0.05}^1$ indicates value clamping.
$\mathcal{R}$ is a random sample index set of $\mathcal{P}$ with $N_{\boldsymbol P}/1.5$ items.
Finally, the sampling probability of a point $p_i \!\in\! \mathcal{P}$ is $\rho(p_i) \!=\! {\upsilon(p_i)}/{\sum_{p_j \in \mathcal{P}} \upsilon(p_j)}$.

\noindent
\textbf{Feature Fusion}.
In order to allow each point in the local patch to have global information and determine the normal orientation, we first use the maxpooling and repetition operation to make the output global latent code has the same dimension as the local latent code.
Then, the two kinds of codes are fused by concatenation, \ie, $[z^{\mathbf{n}}_q : z^s_q]$ in Eq.~\eqref{eq:poly}.

\subsection{Loss Functions}

For the query point $q$, we constrain its unoriented normal and normal sign (\ie, orientation), respectively.
To learn an accurate unoriented normal, we employ the ground truth $\hat{\mathbf{n}}_q$ to calculate a normal vector $sin$ loss~\cite{ben2020deepfit}
\begin{equation}
  \mathcal{L}_{sin} = \|\mathbf{n}_q \times \hat{\mathbf{n}}_q \|.
\end{equation}
For the normal orientation, we adopt the binary cross entropy $H$~\cite{erler2020points2surf} to calculate a sign classification loss
\begin{equation}
  \mathcal{L}_{sgn} = H \Big( \sigma \big( g^s(q)\big),\ [f_{\boldsymbol S}(q) > 0] \Big),
\end{equation}
where $\sigma$ is a logistic function that converts the sign logits to probabilities.
$[f_{\boldsymbol S}(q) \!>\! 0]$ is $1$ if the estimated normal faces the outward of surface ${\boldsymbol S}$ and $0$ otherwise.
Our method achieves a significant performance boost by dividing the oriented normal estimation into unoriented normal regression and its sign classification, instead of directly regressing the oriented normals of query points (see ablations in Sec.~\ref{sec:ablation}).

To facilitate the local feature learning and make the model also pay attention to the orientation consistency of neighboring points $p_i \!\in\! {\boldsymbol p}_q$,
we compute a weighted mean square error (MSE)
\begin{equation} \label{eq:loss_n}
  \mathcal{L}_{mse} \!=\! \frac{1}{N} \sum_{i=1}^{N} \tau_{i} \|\vec{\mathbf{n}}_i - \hat{\vec{\mathbf{n}}}_i \|^2,
\end{equation}
where the neighborhood point normals $\vec{\mathbf{n}} \!=\! \delta({z_{q}})$ are predicted from the surface embedding $z_{q}$ by an MLP layer $\delta : \mathbb{R}^c \!\to\! \mathbb{R}^3$.
Moreover, we add a loss term according to coplanarity~\cite{zhang2022geometry} to facilitate the learning of $\tau$ in Eq.\eqref{eq:output},
\begin{equation}  \label{eq:loss_w}
  \mathcal{L}_{\tau} = \frac{1}{N} \sum_{i=1}^{N}(\tau_{i} - \hat{\tau}_{i})^2,
  ~~ \hat{\tau}_{i} = \exp \left(- \frac{(p_i \cdot \hat{\mathbf{n}}_q)^2}{\xi^2} \right),
\end{equation}
where $\xi \!=\! \max (0.0025, ~0.3 \sum_{i=1}^{N}(p_i \cdot \hat{\mathbf{n}}_q)^2 / N )$.
In summary, our final training loss for oriented normal estimation is
\begin{equation}
  \mathcal{L} = \lambda_1 \mathcal{L}_{sin} + \lambda_2 \mathcal{L}_{sgn} + \lambda_3 \mathcal{L}_{mse} + \lambda_4 \mathcal{L}_{\tau} ~,
\end{equation}
where $\lambda_1\!=\!0.1$, $\lambda_2\!=\!0.1$, $\lambda_3\!=\!0.5$ and $\lambda_4\!=\!1.0$ are weighting factors that are first set empirically and then fine-tuned based on experiments.

\subsection{Unoriented Normal Estimation}  \label{sec:unoriented}

In the previous sections, we introduce to estimate oriented normals by implicitly learning signed hyper surfaces in the feature space.
The network model extracts local and global feature representations through patch and shape encoding respectively, and the global features from shape encoding help determine the normal orientation.
In this section, we show that the modules we designed in the patch encoding can be reorganized to estimate unoriented normals, and their orientations are solved in the next section.

To estimate unoriented normals, \ie, the local property of point clouds, whose orientations are not guaranteed to be globally consistent, the global information from shape encoding is not needed.
Thus, we build an unoriented normal estimation pipeline by using the local latent code extraction layer $\mathcal{F}$ in Eq.~\eqref{eq:local}.
The input of this new network pipeline is the local point cloud patch and its output is the unoriented normal of the query point.
The layer $\mathcal{F}$ is stacked recursively, enabling the network model to learn increasingly rich representations of the point cloud patch.
For that, the features $\mathcal{X}$ from two layers are aggregated by add operation and passed to the next layer, \ie,
\begin{equation}
  \mathcal{X}_{k+1} = [\mathcal{X}_k]_{N_{k+1}} + \mathcal{F}_2(\mathcal{X}_k), ~ \mathcal{X}_{k} = \mathcal{F}_1(\mathcal{X}_{k-1}),
\end{equation}
where $[\cdot]_{N_{k+1}}$ denotes the neighborhood scale of size ${N_{k+1}}$ with respect to query point.
By continuously reducing the number of k-nearest neighbors of the query point, our layers extract features from different scales of the query point in order from large to small.
With the recursive utilization of our layers, the large scales of earlier layers give more robust information about the underlying geometries, while the small scales of the latter layers lead to a more accurate description of the local details.
In this manner, the features from different scales of the local patch around a query point are fused to obtain its optimal geometric description.

After obtaining the final output feature $\mathcal{X}_o$ with $N_o$ neighboring points in the patch, the unnormalized normal $\mathbf{n}_q$ of the query point is predicted by a weighted maxpooling of its neighboring features $x_{i} \!\in\! \mathcal{X}_o$, that is
\begin{equation} \label{eq:output_uo}
  \mathbf{n}_q = \mathcal{O} \big(\text{MAX} \{ w_i \cdot \tau_i \cdot x_{i} | i\!=\!1,\cdots,N_o \} \big) ,
\end{equation}
where $\tau_i \!=\! \text{sigmoid}(\mathcal{I}(x_{i}))$ is the point weight. $\mathcal{O}$ and $\mathcal{I}$ are MLPs.
Furthermore, the neighboring point normals $\mathbf{n}_{i}$ are predicted from $\mathcal{X}_o$ by another MLP.

\noindent
\textbf{Training Loss}.
To constrain the predicted normal of the query point, we calculate the $sin$ distance $d_{\rm sin}$ and squared Euclidean distance $d_{\rm euc}$ between the predicted normal $\mathbf{n}$ and the ground truth normal $\hat{\mathbf{n}}$, \ie,
\begin{equation}  \label{eq:loss_uo}
  \mathcal{L}_q = \|\mathbf{n} \times \hat{\mathbf{n}} \|
                      + \text{min}\left(\|\mathbf{n} - \hat{\mathbf{n}}\|^2, ~~ \|\mathbf{n} + \hat{\mathbf{n}}\|^2 \right).
\end{equation}
Meanwhile, we calculate a weighted neighborhood consistency loss based on the ground truth normals $\hat{\mathbf{n}}_{i}$ of neighboring points, then we have
\begin{equation}
  \mathcal{L}_{con} \!=\! \frac{1}{N_o} \sum_{i=1}^{N_o} \tau_{i} \left(\|\mathbf{n}_{i} \times \hat{\mathbf{n}}_{i} \|
                      + \text{min}\left(|\mathbf{n}_i - \hat{\mathbf{n}}_i|^2, |\mathbf{n}_i + \hat{\mathbf{n}}_i|^2 \right)\right).
\end{equation}
Thus, we obtain the loss $\mathcal{L}_q$ for query point normal $\mathbf{n}_q$ and the mean loss $\mathcal{L}_{con}$ for neighboring point normals $\mathbf{n}_{i}$.
The distance-based weight $w$ in Eq.~\eqref{eq:weight} and the loss function of $\tau$ in Eq.~\eqref{eq:loss_w} are also adopted in this section.

In summary, the final training loss function for unoriented normal estimation is given by
\begin{equation}  \label{eq:finalloss}
  \setlength{\abovedisplayskip}{6pt}
  \setlength{\belowdisplayskip}{6pt}
  \mathcal{L} = \kappa_1 \mathcal{L}_q + \kappa_2 \mathcal{L}_{con} + \kappa_3 \mathcal{L}_{\tau},
\end{equation}
where the weighting factors are set to $\kappa_1\!=\!0.1$, $\kappa_2\!=\!0.4$ and $\kappa_3\!=\!1.0$.
We first select their initial values empirically and then fine-tune the parameters experimentally.

\begin{table*}[t]
	\centering
	\footnotesize
	\setlength{\tabcolsep}{1.5mm}
	\caption{
		Unoriented normal evaluation on datasets PCPNet and FamousShape.
		We report RMSE results under different noise levels and different sampling ways, and rank the methods according to the average RMSE on the PCPNet dataset.
        $\ast$ means the code is uncompleted or unavailable, and the result comes from its public paper.
		'Ours' denotes the result of our oriented normals and 'Ours-U' denotes the result of our unoriented normals.
    }
	\vspace{-0.4cm}
    \label{table:pcpnet_famousShape}
    \resizebox{\linewidth}{!}{
	\begin{tabular}{l|cccc|cc| >{\columncolor{mygray}} c||  cccc|cc| >{\columncolor{mygray}} c}
		\toprule
		\multirow{3}{*}{Category} & \multicolumn{7}{c||}{\textbf{PCPNet Dataset}} & \multicolumn{7}{c}{\textbf{FamousShape Dataset}} \\
		\cmidrule(r){2-15}
		& \multicolumn{4}{c|}{Noise} & \multicolumn{2}{c|}{Density} &     & \multicolumn{4}{c|}{Noise} & \multicolumn{2}{c|}{Density} &    \\
		& None & 0.12\% & 0.6\% & 1.2\%  & Stripe & Gradient  & \multirow{-2}{*}{{Average}} & None & 0.12\% & 0.6\% & 1.2\%  & Stripe & Gradient & \multirow{-2}{*}{{Average}} \\
		\midrule
		Jet~\cite{cazals2005estimating}              & 12.35 & 12.84 & 18.33 & 27.68 & 13.39 & 13.13 &  16.29    & 20.11 & 20.57 & 31.34 & 45.19 & 18.82 & 18.69 &   25.79  \\
		PCA~\cite{hoppe1992surface} 	             & 12.29 & 12.87 & 18.38 & 27.52 & 13.66 & 12.81 &  16.25    & 19.90 & 20.60 & 31.33 & 45.00 & 19.84 & 18.54 &   25.87  \\
		PCPNet~\cite{guerrero2018pcpnet}             & 9.64  & 11.51 & 18.27 & 22.84 & 11.73 & 13.46 &  14.58    & 18.47 & 21.07 & 32.60 & 39.93 & 18.14 & 19.50 &   24.95  \\
        Zhou \etal$^\ast$~\cite{zhou2020normal}      & 8.67  & 10.49 & 17.62 & 24.14 & 10.29 & 10.66 &  13.62    & -     & -     & -     & -     & -     & -     & -        \\
		Nesti-Net~\cite{ben2019nesti}                & 7.06  & 10.24 & 17.77 & 22.31 & 8.64  & 8.95  &  12.49    & 11.60 & 16.80 & 31.61 & 39.22 & 12.33 & 11.77 &   20.55  \\
		Lenssen \etal~\cite{lenssen2020deep}         & 6.72  & 9.95  & 17.18 & 21.96 & 7.73  & 7.51  &  11.84    & 11.62 & 16.97 & 30.62 & 39.43 & 11.21 & 10.76 &   20.10  \\
		DeepFit~\cite{ben2020deepfit}                & 6.51  & 9.21  & 16.73 & 23.12 & 7.92  & 7.31  &  11.80    & 11.21 & 16.39 & 29.84 & 39.95 & 11.84 & 10.54 &   19.96  \\
        MTRNet$^\ast$~\cite{cao2021latent}           & 6.43  & 9.69  & 17.08 & 22.23 & 8.39  & 6.89  &  11.78    & -     & -     & -     & -     & -     & -     & -        \\
        Refine-Net~\cite{zhou2022refine}             & 5.92  & 9.04  & 16.52 & 22.19 & 7.70  & 7.20  &  11.43    & -     & -     & -     & -     & -     & -     & -        \\
        Zhang \etal$^\ast$~\cite{zhang2022geometry}  & 5.65  & 9.19  & 16.78 & 22.93 & 6.68  & 6.29  &  11.25    & 9.83  & 16.13 & 29.81 & 39.81 & 9.72  & 9.19  &   19.08  \\
        Zhou \etal$^\ast$~\cite{zhou2023improvement} & 5.90  & 9.10  & 16.50 & 22.08 & 6.79  & 6.40  &  11.13    & -     & -     & -     & -     & -     & -     & -        \\
        AdaFit~\cite{zhu2021adafit}                  & 5.19  & 9.05  & 16.45 & 21.94 & 6.01  & 5.90  &  10.76    & 9.09  & 15.78 & 29.78 & 38.74 & 8.52  & 8.57  &   18.41  \\
		GraphFit~\cite{li2022graphfit}               & 5.21  & 8.96  & 16.12 & 21.71 & 6.30  & 5.86  &  10.69    & 8.91  & 15.73 & 29.37 & 38.67 & 9.10  & 8.62  &   18.40  \\
		NeAF~\cite{li2023NeAF}                       & 4.20  & 9.25  & 16.35 & 21.74 & 4.89  & 4.88  &  10.22    & 7.67  & 15.67 & 29.75 & 38.76 & 7.22  & 7.47  &   17.76  \\
		HSurf-Net~\cite{li2022hsurf}                 & 4.17  & 8.78  & 16.25 & 21.61 & 4.98  & 4.86  &  10.11    & 7.59  & 15.64 & 29.43 & 38.54 & 7.63  & 7.40  &   17.70  \\
		NGLO~\cite{li2023NGLO}  	                 & 4.06  & 8.70  & 16.12 & 21.65 & 4.80  & 4.56  &  9.98     & 7.25  & 15.60 & 29.35 & 38.74 & 7.60  & 7.20  &   17.62  \\
		Du \etal~\cite{du2023rethinking}             & 3.85  & 8.67  & 16.11 & 21.75 & 4.78  & 4.63  &  9.96     & 6.92  & 15.05 & 29.49 & 38.73 & \textbf{7.19}  & 6.92  &   17.38  \\
		CMG-Net~\cite{wu2024cmg}                     & 3.87  & 8.45  & 16.08 & 21.89 & 4.85  & 4.45  &  9.93     & 7.07  & 14.83 & 29.04 & 38.93 & 7.43  & 7.03  &   17.39  \\
		MSECNet~\cite{xiu2023msecnet}                & 3.84  & 8.74  & 16.10 & \textbf{21.05} & \textbf{4.34}  & 4.51  &  9.76     & -     & -     & -     & -     & -     & -     & -        \\
		Ours    	                                 & 3.95  & 8.55  & 16.13 & 21.53 & 4.91  & 4.67  &  9.96     & 7.41  & 15.34 & 29.33 & 38.56 & 7.74  & 7.28  &   17.61  \\
		Ours-U      & \textbf{3.49} & \textbf{8.43}  & \textbf{15.73} & \textbf{21.05} &  {4.45} & \textbf{4.17} &  \textbf{9.55}
					& \textbf{6.79} & \textbf{15.04} & \textbf{29.03} & \textbf{38.40} &  {7.21} & \textbf{6.67} &  \textbf{17.19}  \\
		\bottomrule
	\end{tabular} }
    \vspace{-0.2cm}
\end{table*}

\subsection{Unoriented Normal Orientation}

Another issue is that the global consistency of the estimated unoriented normals in Sec.~\ref{sec:unoriented} cannot be guaranteed since they are obtained by regression from local features only.
We know that unoriented normal is a local property of the point cloud, while oriented normal estimation requires additional information to determine its orientation.
Thus, the unoriented normal estimation in Sec.~\ref{sec:unoriented} solely uses the local information, and the oriented normal estimation in Sec.~\ref{sec:nor_esti} further incorporates global information to learn signed hyper surfaces to predict the normal and its sign.
Here we show that we can tune the orientation of unoriented normal $\mathbf{n}$ to achieve its consistent direction by using the oriented normal result $\vec{\mathbf{n}}$ as the reference.
For this, we introduce a normal orientation strategy to transfer the normal sign of oriented normal to the unoriented normal and re-orientate its direction.
Different from the local orientation propagation using MST~\cite{hoppe1992surface}, our strategy is based on the angle distance between the corresponding oriented and unoriented normal vectors at the same point, and the new oriented normal $\vec{\mathbf{n}}'$ is obtained by
\begin{equation}
  \vec{\mathbf{n}}' = \left\{
    \begin{aligned}
       \mathbf{n},    ~~~~& {\rm if}~~ \mathbf{n} \cdot \vec{\mathbf{n}} > 0~; \\
      -\mathbf{n},    ~~~~& {\rm if}~~ \mathbf{n} \cdot \vec{\mathbf{n}} < 0~,
    \end{aligned}
  \right.
\end{equation}
where the reference normal $\vec{\mathbf{n}}$ can be the oriented normal estimated in this work or obtained through other methods.
We will experimentally show that we can employ the unoriented normals to further improve the accuracy of oriented normal estimation results.

\begin{figure}[t]
  \centering
  \includegraphics[width=\linewidth]{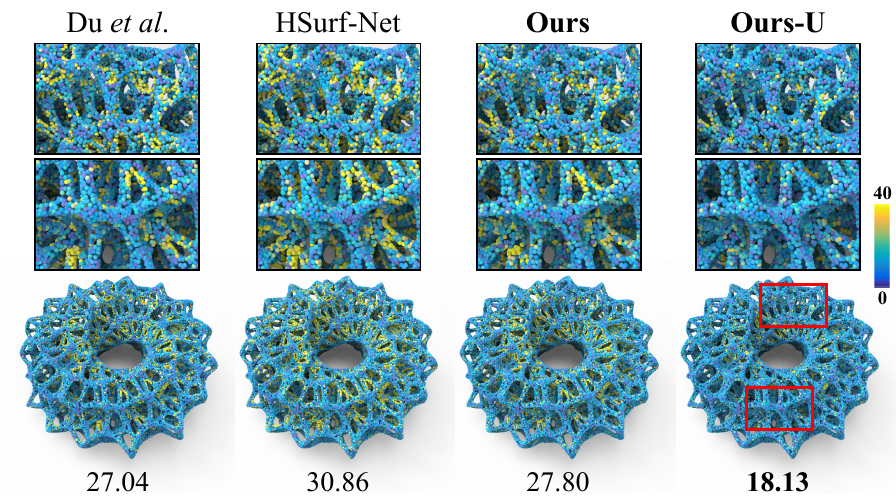}  \vspace{-0.8cm}
  \caption{
    Visual comparison of unoriented normal errors on a point cloud with complex geometry.
    The normal RMSE is mapped to a heatmap ($0^{\circ}-40^{\circ}$).
    We provide the average RMSE over shape for each method.
  }
  \label{fig:errorMap_NestPC}
  \vspace{-0.2cm}
\end{figure}

\section{Experiments}

\noindent
\textbf{Implementation}.
We only train our network model on the PCPNet shape dataset~\cite{guerrero2018pcpnet}, which provides the ground truth normals with consistent orientation (outward of the surface).
We follow the same train/test data split and data processing as in~\cite{guerrero2018pcpnet,ben2020deepfit,zhu2021adafit,li2022hsurf}.
For patch encoding, we randomly select a query point from the shape point cloud and search its $700$ neighbors to form a patch.
For shape encoding, we sample $N_{\boldsymbol P} \!=\! 1200$ points from the shape point cloud according to the sampling probability.
The Adam optimizer is adopted with an initial learning rate of $9 \!\times\! 10^{-4}$ which is decayed to $1/5$ of the latest value at epochs $\{400,600,800\}$.
The model is trained on an NVIDIA 2080 Ti GPU with a batch size of $145$ and epochs of $800$.

\noindent
\textbf{FamousShape Dataset}.
Due to the lack of relevant datasets and the relatively simple test shapes of the PCPNet dataset~\cite{guerrero2018pcpnet},
we further collect shapes with complex structures from other public datasets, such as the Famous dataset~\cite{erler2020points2surf} and the Stanford 3D Scanning Repository~\cite{curless1996volumetric}.
We sample 100K points from each shape and follow the same preprocessing steps as the PCPNet dataset to conduct data augmentation, \eg, adding Gaussian noise with different levels (0.12\%, 0.6\% and 1.2\%) and non-uniform sampling (stripe and gradient).
The ground truths of oriented normals are extracted from mesh data and used for evaluation.
We call this dataset \emph{FamousShape}, and it is available along with our code.
In the supplementary material, we visualize the point cloud shapes of the FamousShape dataset, which has more complex geometries than the PCPNet dataset.

\noindent
\textbf{Evaluation Metrics}.
We adopt the same evaluation metrics as in~\cite{guerrero2018pcpnet,ben2020deepfit,zhu2021adafit,li2022hsurf} to evaluate the estimated normals.
More specifically, Root Mean Squared Error (RMSE) measures normal angles between the ground truth normals $\hat{\mathbf{n}}$ and the predicted normals $\mathbf{n}$, while the curve of Percentage of Good Point (PGP) shows the overall quality of results by counting points whose normal errors are less than the given thresholds.
They are computed by
\begin{align}
  {\rm RMSE} &= \sqrt{\frac{1}{N} \sum_{i=1}^{N} \big( \arccos (\phi) \big)^2} ~~, \\
  {\rm PGP}(\tau) &= \frac{1}{N} \sum_{{i}=1}^{N} \mathcal{I} \big( \arccos (\phi) < \tau \big) ~~, 
\end{align}
where $N$ is the number of evaluated normals in a point cloud.
$\phi$ is the angle cosine of two vectors, and $\phi_{\rm unoriented} \!=\! |\hat{\mathbf{n}}_i \odot \mathbf{n}_i |$ and $\phi_{\rm oriented} \!=\! \hat{\mathbf{n}}_i \odot \mathbf{n}_i$ are used in unoriented and oriented normal evaluation, respectively.
$|\cdot|$ represents the absolute value of the inner product $\odot$ of two normal vectors.
Therefore, the normal angle error ${\rm RMSE}_{\rm unoriented}$ is bounded between $0^{\circ}$ and $90^{\circ}$ in unoriented normal evaluation, and ${\rm RMSE}_{\rm oriented}$ is bounded between $0^{\circ}$ and $180^{\circ}$ in oriented normal evaluation.
$\mathcal{I}$ represents an indicator function that measures whether the error is less than a given threshold $\tau$.
The ground truth normals in the benchmark datasets face outward of the shape surface.
For the baseline methods, we flip their estimated normals if more than half of the normals face inward during oriented normal evaluation.

\emph{
In the following experiments, we use 'Ours' to denote the result of our oriented normals, 'Ours-U' to denote the result of our unoriented normals, 'Ours-U+O' to denote the result of our unoriented normals being re-orientated by our oriented normal, and 'Ours-U+NGL' to denote the result of our unoriented normals being re-orientated using the NGL module~\cite{li2023NGLO}.}

\begin{table}[t]
    \footnotesize
    \centering
    \setlength{\tabcolsep}{0.5mm}
    \caption{
      Comparison of unoriented normal PGP($20^\circ$) on the datasets PCPNet and FamousShape under the highest noise.
      The higher the better.
    }
    \vspace{-0.4cm}
    \resizebox{\linewidth}{!}{
    \begin{tabular}{c|ccccccccc}
      \toprule
      (\%)          & GraphFit & NeAF   & HSurf-Net  & NGLO   & Du \etal  & CMG-Net  & Ours    &  Ours-U           \\  
      \midrule
      PCPNet        & 77.71    & 77.44  & 77.77      & 77.76  & 77.79     & 77.35    & \textbf{77.94}   & {77.80}  \\  
      Famous.       & 43.99    & 44.05  & 44.62      & 44.19  & 43.97     & 43.18    & 44.67   & \textbf{44.69}    \\  
      \bottomrule
    \end{tabular} }
    \label{tab:PGP}
    \vspace{-0.2cm}
\end{table}

\begin{table}[t]
	\centering
	\footnotesize
	\setlength{\tabcolsep}{.5mm}
	\caption{
		Unoriented normal RMSE on the datasets SceneNN and ScanNet.
    }
	\vspace{-0.35cm}
    \label{table:scenenn}
    \resizebox{\linewidth}{!}{
	\begin{tabular}{c|cccccccc}
		\toprule
		            & GraphFit & NeAF  & HSurf-Net & NGLO   & Du \etal & CMG-Net  & Ours   & Ours-U           \\  
		\midrule
		SceneNN     & 8.59     & 7.88  & 7.55      & 7.73   & 7.68     & 7.64     & 7.93   & \textbf{7.30}    \\  
		ScanNet     & 18.31    & 15.48 & 15.83     & 15.44  & 15.37    & 15.14    & 15.75  & \textbf{15.02}   \\  
		\bottomrule
	\end{tabular} }
    \vspace{-0.2cm}
\end{table}

\begin{figure}[t]
  \centering
  \includegraphics[width=\linewidth]{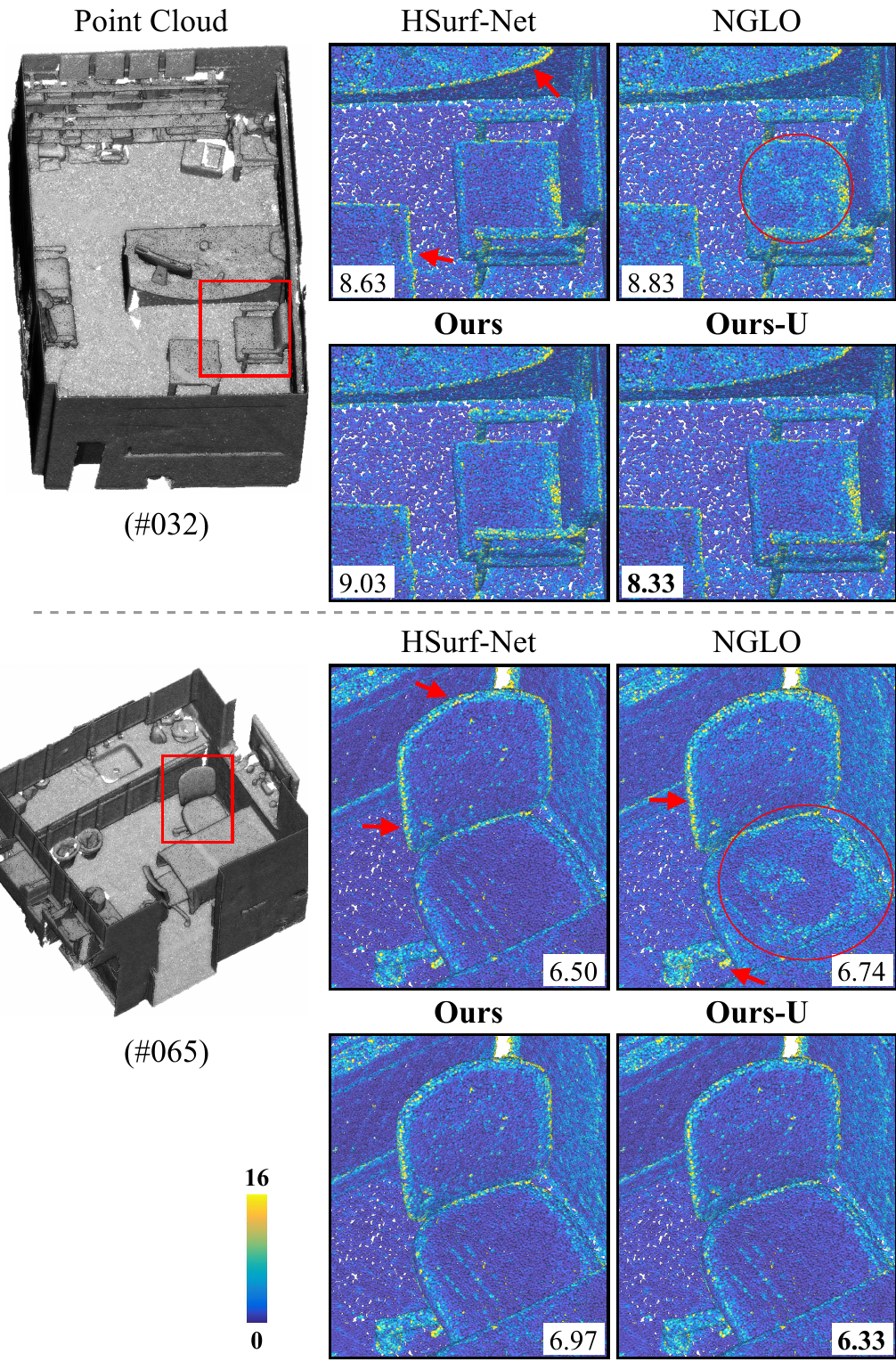}  \vspace{-0.7cm}
  \caption{
    Visual comparison of unoriented normal errors on the SceneNN dataset.
    The normal RMSE is mapped to a heatmap ($0^{\circ}-16^{\circ}$).
    We provide the average RMSE of each method over the entire point cloud.
  }
  \label{fig:errorMap_SceneNN}
  \vspace{-0.2cm}
\end{figure}

\subsection{Unoriented Normal Comparison}

We use our estimation results of \emph{oriented} and \emph{unoriented} normal to compare with baseline methods that are designed for estimating \emph{unoriented} normals, such as the traditional methods PCA~\cite{hoppe1992surface} and Jet~\cite{cazals2005estimating}, the learning-based surface fitting methods DeepFit~\cite{ben2020deepfit}, AdaFit~\cite{zhu2021adafit} and GraphFit~\cite{li2022graphfit}, and the learning-based regression methods PCPNet~\cite{guerrero2018pcpnet}, Nesti-Net~\cite{ben2019nesti} and HSurf-Net~\cite{li2022hsurf}.
As shown in Table~\ref{table:pcpnet_famousShape}, we report quantitative comparison results with the baselines in terms of normal angle RMSE on two datasets, PCPNet and FamousShape.
On the PCPNet dataset, our method achieves the best performance under almost all noise levels and density variations.
On our FamousShape dataset, our method achieves the best performance under most metrics and has the lowest average RMSE result.
In Fig.~\ref{fig:errorMap_NestPC}, we provide visual comparisons of the unoriented normal error of various methods, and the results show that our method can handle complex geometries better.
In Table~\ref{tab:PGP}, we use the metric of PGP($20^\circ$) to quantitatively evaluate the accuracy of normal results, \ie, the percentage of points whose normal errors are less than $20^\circ$.
The evaluation results show that our method can obtain accurate normals for more points than baseline methods.

\begin{figure}[t]
  \centering
  \includegraphics[width=\linewidth]{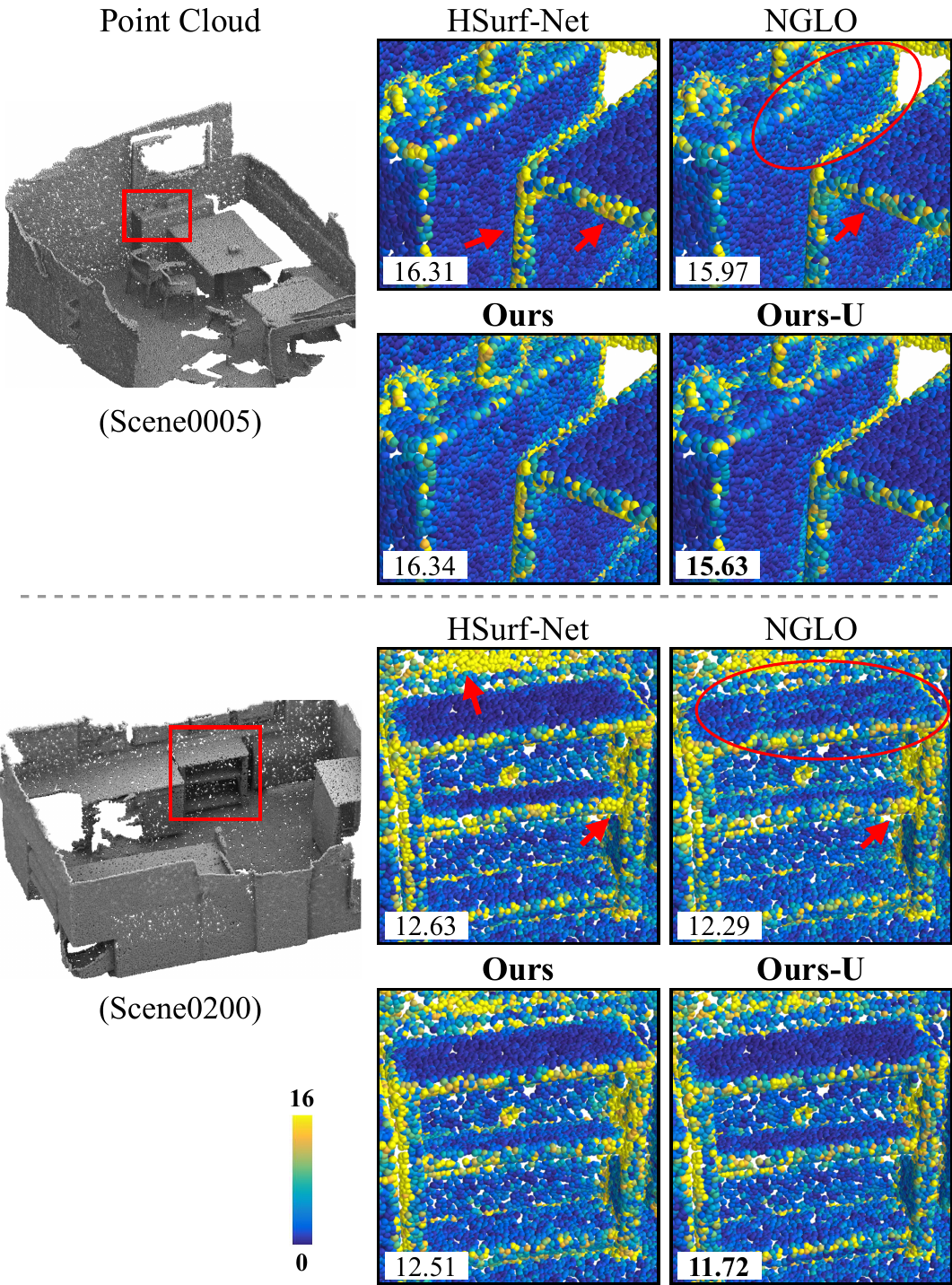}  \vspace{-0.7cm}
  \caption{
    Visual comparison of unoriented normal errors on the ScanNet dataset.
    The normal RMSE is mapped to a heatmap ($0^{\circ}-16^{\circ}$).
    We provide the average RMSE of each method over the entire point cloud.
  }
  \label{fig:errorMap_ScanNet}
  \vspace{-0.2cm}
\end{figure}

\noindent\textbf{Evaluation on Scene Point Clouds}.
To evaluate the generalization ability of our method, we use the network models (including models for oriented and unoriented normal estimation) trained on the PCPNet shape dataset to test on real-scanned scene data of datasets SceneNN~\cite{hua2016scenenn} and ScanNet~\cite{dai2017scannet}.
For these two indoor scene datasets, we only report the quantitative evaluation results for unoriented normals, rather than oriented normals, because it is ambiguous to judge the internal or external orientation of normals of objects and walls inside a room.
Unless these objects and walls are segmented very precisely, which is not easy to achieve.
The ground truth normal is obtained from the provided mesh data.
In Table~\ref{table:scenenn}, we provide the evaluation results of unoriented normals on these two datasets, and our method achieves significant improvements compared to baseline methods.
In Fig.~\ref{fig:errorMap_SceneNN} and Fig.~\ref{fig:errorMap_ScanNet}, we provide visual comparisons of the unoriented normal error on some indoor room scenes of datasets, SceneNN and ScanNet, respectively.
These comparison results demonstrate the good generalization ability and outstanding performance of our method.
In the supplementary material, we provide more visual comparisons of the unoriented normal error on real-scanned indoor scenes.

\begin{figure*}[t]
  \centering
  \subfigure[PCPNet dataset]{
    \includegraphics[width=.48\linewidth]{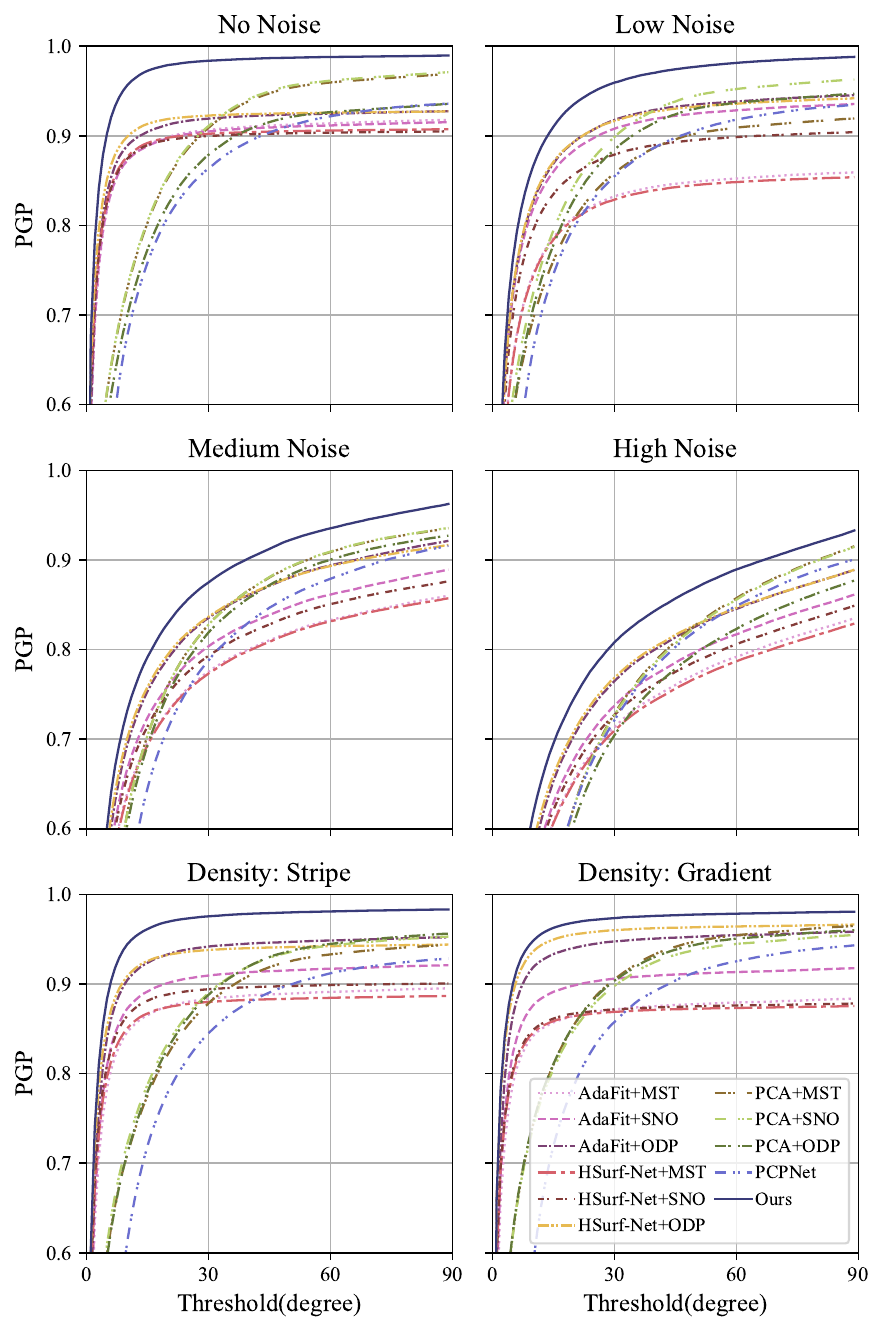}  
  }
  \subfigure[FamousShape dataset]{
    \includegraphics[width=.48\linewidth]{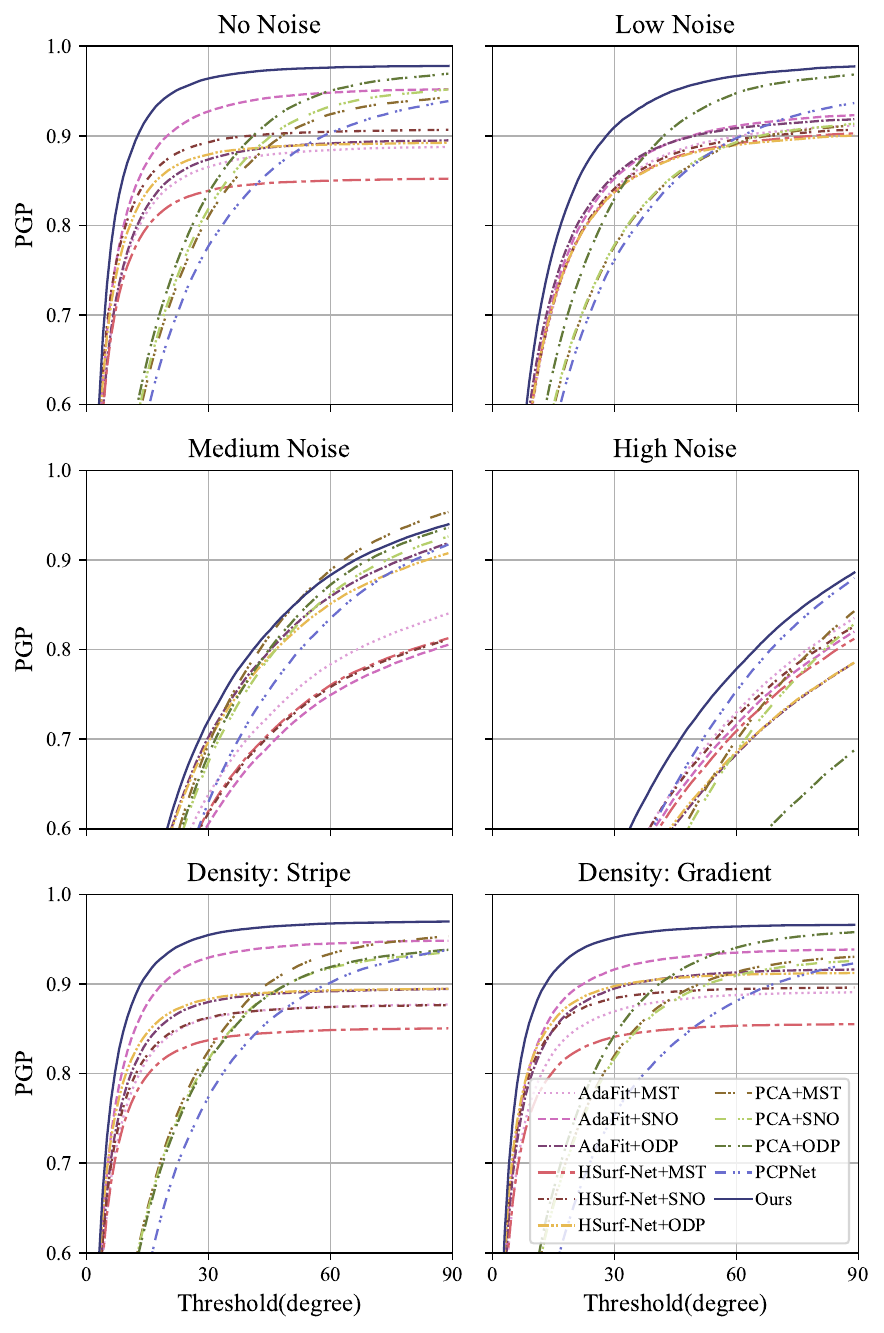}  
  }
  \vspace{-0.3cm}
  \caption{
    PGP of oriented normals on the PCPNet dataset and the FamousShape dataset.
    It shows the percentage of correctly estimated point normals for a given angle threshold.
    Our method has the best PGP result for almost all thresholds.
  }
  \label{fig:PGP}
  \vspace{-0.2cm}
\end{figure*}

\subsection{Oriented Normal Comparison}

\begin{figure}[t]
  \centering
  \includegraphics[width=\linewidth]{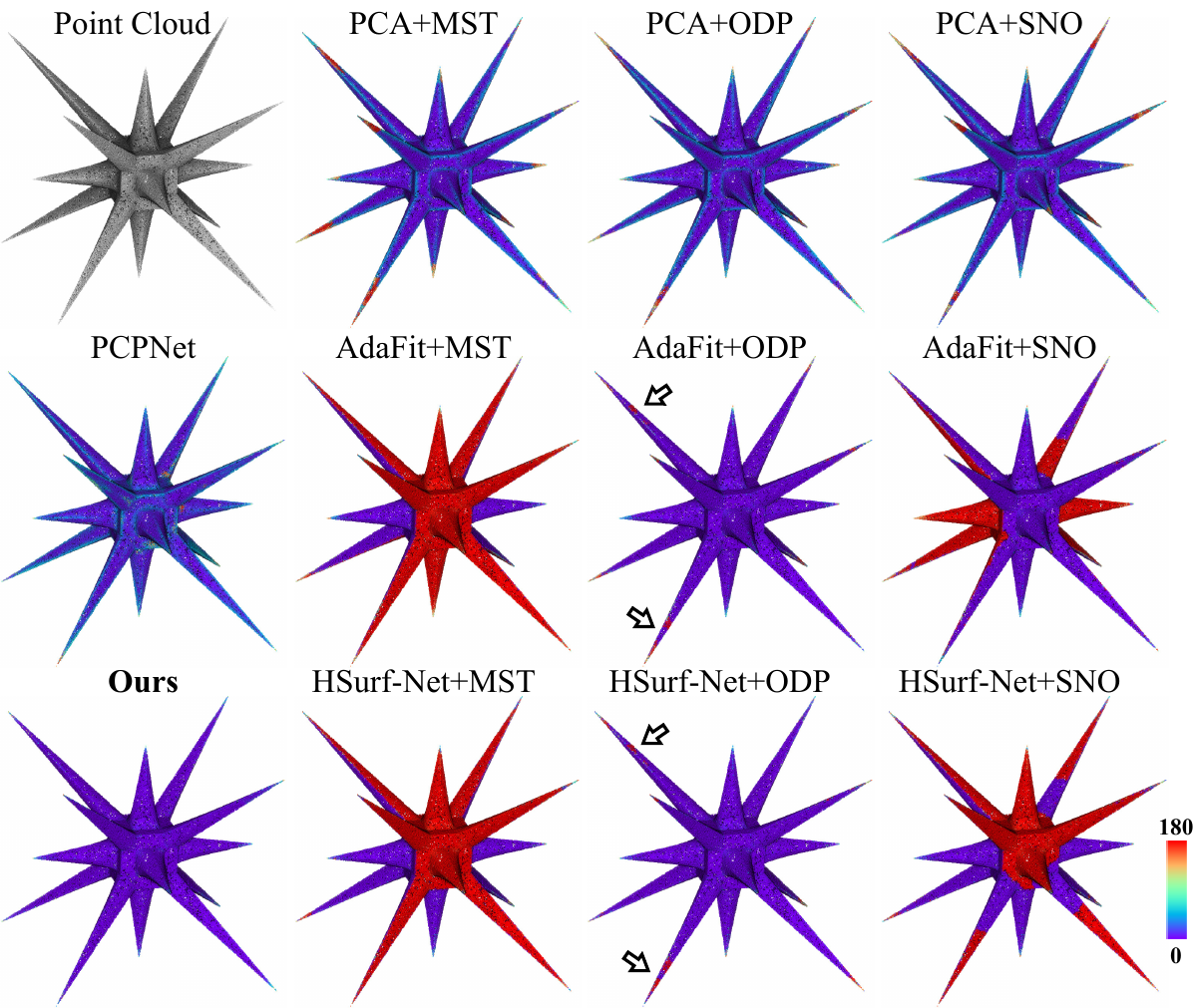}  \vspace{-0.85cm}
  \caption{
    Visualization of the oriented normal error on a point cloud with sharp features.
    The angle error is mapped to a heatmap ranging from $0^{\circ}$ to $180^{\circ}$.
    Our method has less error than other baseline methods.
  }
  \label{fig:errorMap_star}
  \vspace{-0.2cm}
\end{figure}

\begin{figure}[t]
  \centering
  \includegraphics[width=\linewidth]{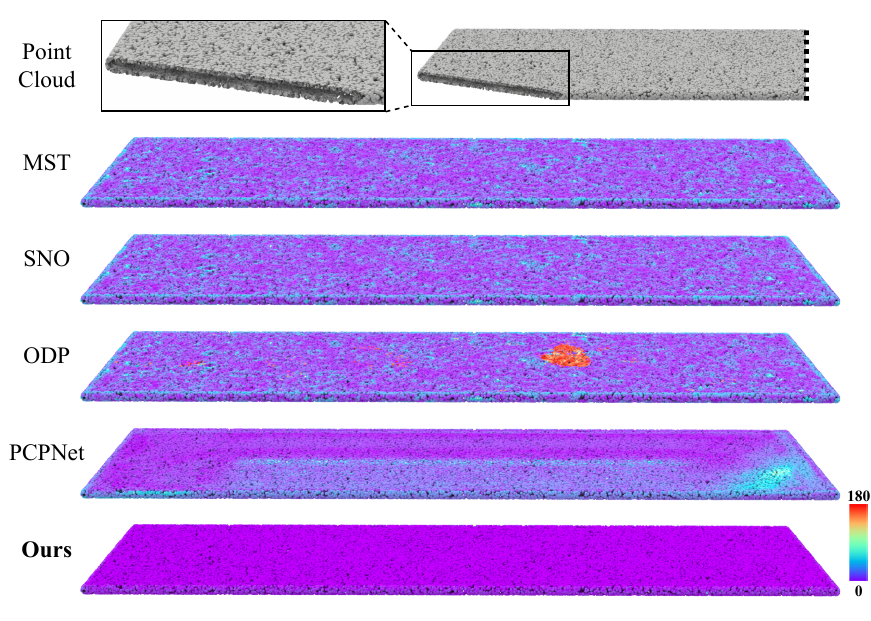}  \vspace{-0.8cm}
  \caption{
    Visualization of the oriented normal error on a thin sheet with a hollow structure.
    The initial unoriented normals for methods MST~\cite{hoppe1992surface}, SNO~\cite{schertler2017towards} and ODP~\cite{metzer2021orienting} are provided by PCA.
    MST and SNO have very similar orientation results, and our method has the least error.
  }
  \label{fig:errorMap_sheet}
\end{figure}

We compare our approach for \emph{oriented} normal estimation with various baseline methods, such as PCPNet~\cite{guerrero2018pcpnet}, DPGO~\cite{wang2022deep} and NGLO~\cite{li2023NGLO}.
The trained model of PCPNet is available.
The source code of DPGO is uncompleted and its results on the PCPNet dataset are taken from its paper.
In addition, we choose three \emph{unoriented} normal estimation methods (PCA~\cite{hoppe1992surface}, AdaFit~\cite{zhu2021adafit} and HSurf-Net~\cite{li2022hsurf}) and three normal orientation methods (MST~\cite{hoppe1992surface}, SNO~\cite{schertler2017towards} and ODP~\cite{metzer2021orienting}), and make different combinations of them to form two-stage pipelines for estimating oriented normals, such as PCA+MST and HSurf-Net+ODP.
Among the baselines, PCA is a widely used traditional method, AdaFit is a representative surface fitting-based method, and HSurf-Net is a regression-based method and has the state-of-the-art performance for \emph{unoriented} normal estimation.
We use the original implementation of SNO and ODP, and the implementation of MST in~\cite{cloudcompare}.
In Table~\ref{table:pcpnet_famousShape_o}, we show the quantitative comparison results on datasets PCPNet and FamousShape.
We can see that our method provides the most accurate normals under almost all noise levels and density variations for both datasets, and achieves huge performance gains in terms of average results compared to all baselines.
From the experimental results, we find that the propagation-based normal orientation methods have significantly varied results when dealing with unoriented normal inputs from different estimation methods, such as PCA+MST and AdaFit+MST.
The overall error distributions of various methods on datasets PCPNet and FamousShape are illustrated in Fig.~\ref{fig:PGP}, and our method achieves excellent performance at different thresholds.
A visual comparison result of the normal errors on a point cloud with sharp corners is shown in Fig.~\ref{fig:errorMap_star}, which shows the superior capability of our method.
As shown in Fig.~\ref{fig:errorMap_sheet}, we provide an example of a point cloud sampled from a thin sheet.
It has two planes that are very close together, which can easily affect the accuracy of the unoriented normal and the orientation of the oriented normal, \ie, which side of the sheet it points to.
The quantitative and qualitative comparison results show that our method has a huge advantage over the baseline methods in the oriented normal estimation task.

\begin{figure}[t]
  \centering
  \includegraphics[width=\linewidth]{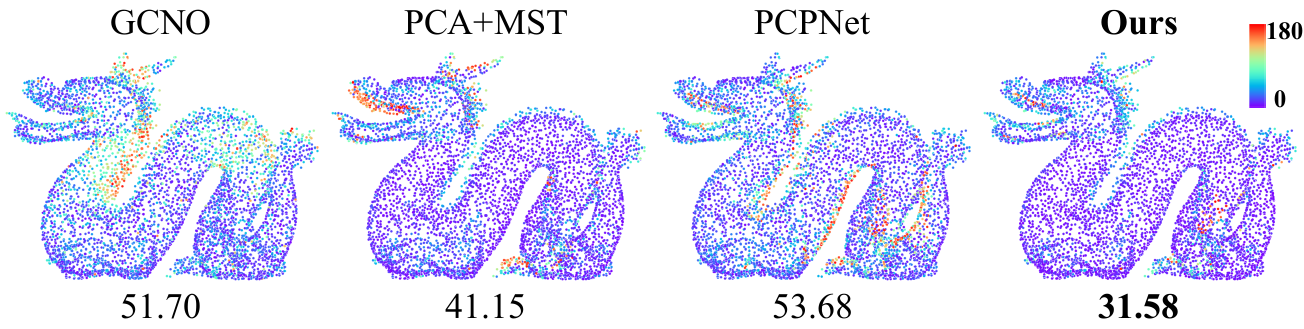}  \vspace{-0.7cm}
  \caption{
    Visualization of the oriented normal error on a sparse point cloud with 5000 points.
    The RMSE of the estimated normal is provided for quantitative comparison, and our method has the lowest error.
  }
  \label{fig:errorMap_sparse}
\end{figure}

\noindent\textbf{Evaluation on Sparse Point Clouds}.
To evaluate the generalization ability of our method, we conduct evaluations on two sets of point clouds that have the same shapes as the FamousShape dataset but each shape in these two sets contains only 3000 and 5000 points, respectively.
We first evaluate our method for unoriented normal estimation.
As shown in Table~\ref{table:sparse}, we report quantitative comparison results of unoriented normals, and our method achieves significant performance improvements.
Then, we evaluate for oriented normal estimation.
The recently proposed algorithm, GCNO~\cite{xu2023globally}, can estimate point cloud normals with globally consistent orientations.
The problem is that its running time increases so drastically with the number of points in the point cloud that we cannot fully test it on existing benchmark datasets, such as PCPNet and FamousShape where each shape has 100K points.
Therefore, in order to complete the evaluation in a reasonable amount of time, we compare with GCNO on sparse data.
As shown in Table~\ref{tab:sparse_o}, we report quantitative comparison results of oriented normal estimation on sparse point clouds.
The traditional baseline algorithms, including GCNO and PCA+MST, are implemented in C++ on the Windows platform and run on an Intel i9-11900K CPU.
And other learning-based methods are implemented in Pytorch on the Linux platform and run on the GPU.
From Table~\ref{tab:sparse_o}, we can see that our method has the best RMSE result and good execution efficiency.
The high time consumption of GCNO limits its application to data with a large number of points.
Furthermore, we show a visual comparison of oriented normal errors in Fig.~\ref{fig:errorMap_sparse}.
These evaluation experimental results demonstrate the excellent performance of our method on sparse point cloud data.

\begin{table*}[t]
	\centering
	\footnotesize
	\setlength{\tabcolsep}{1.3mm}
	\caption{
		Oriented normal evaluation results on datasets PCPNet and FamousShape.
		Among the baseline methods, there are schemes that directly estimate the oriented normal, as well as schemes based on two stages.
		$\ast$ means the source code is uncompleted and the result is from its paper.
		'Ours' denotes the result of our oriented normals, 'Ours-U+O' denotes the result of our unoriented normals being re-orientated by our oriented normal, and 'Ours-U+NGL' denotes the result of our unoriented normals being re-orientated using the NGL module~\cite{li2023NGLO}.
    }
	\vspace{-0.4cm}
    \label{table:pcpnet_famousShape_o}
    \resizebox{\linewidth}{!}{
	\begin{tabular}{l|cccc|cc| >{\columncolor{mygray}} c||  cccc|cc| >{\columncolor{mygray}} c}
		\toprule
		\multirow{3}{*}{Category} & \multicolumn{7}{c||}{\textbf{PCPNet Dataset}} & \multicolumn{7}{c}{\textbf{FamousShape Dataset}} \\
		\cmidrule(r){2-15}
		& \multicolumn{4}{c|}{Noise} & \multicolumn{2}{c|}{Density} &     & \multicolumn{4}{c|}{Noise} & \multicolumn{2}{c|}{Density} &    \\
		& None & 0.12\% & 0.6\% & 1.2\%  & Stripe & Gradient  & \multirow{-2}{*}{{Average}} & None & 0.12\% & 0.6\% & 1.2\%  & Stripe & Gradient & \multirow{-2}{*}{{Average}} \\
		\midrule
		PCA~\cite{hoppe1992surface}+MST~\cite{hoppe1992surface}
		& 19.05 & 30.20 & 31.76 & 39.64 & 27.11 & 23.38 &   28.52    & 35.88 & 41.67 & \textbf{38.09} & 60.16 & 31.69 & 35.40 &   40.48  \\
		PCA~\cite{hoppe1992surface}+SNO~\cite{schertler2017towards}
		& 18.55 & 21.61 & 30.94 & 39.54 & 23.00 & 25.46 &   26.52    & 32.25 & 39.39 & 41.80 & 61.91 & 36.69 & 35.82 &   41.31  \\
		PCA~\cite{hoppe1992surface}+ODP~\cite{metzer2021orienting}
		& 28.96 & 25.86 & 34.91 & 51.52 & 28.70 & 23.00 &   32.16    & 30.47 & 31.29 & 41.65 & 84.00 & 39.41 & 30.72 &   42.92  \\
		AdaFit~\cite{zhu2021adafit}+MST~\cite{hoppe1992surface}
		& 27.67 & 43.69 & 48.83 & 54.39 & 36.18 & 40.46 &   41.87    & 43.12 & 39.33 & 62.28 & 60.27 & 45.57 & 42.00 &   48.76  \\
		AdaFit~\cite{zhu2021adafit}+SNO~\cite{schertler2017towards}
		& 26.41 & 24.17 & 40.31 & 48.76 & 27.74 & 31.56 &   33.16    & 27.55 & 37.60 & 69.56 & 62.77 & 27.86 & 29.19 &   42.42  \\
		AdaFit~\cite{zhu2021adafit}+ODP~\cite{metzer2021orienting}
		& 26.37 & 24.86 & 35.44 & 51.88 & 26.45 & 20.57 &   30.93    & 41.75 & 39.19 & 44.31 & 72.91 & 45.09 & 42.37 &   47.60  \\
		HSurf-Net~\cite{li2022hsurf}+MST~\cite{hoppe1992surface}
		& 29.82 & 44.49 & 50.47 & 55.47 & 40.54 & 43.15 &   43.99    & 54.02 & 42.67 & 68.37 & 65.91 & 52.52 & 53.96 &   56.24  \\
		HSurf-Net~\cite{li2022hsurf}+SNO~\cite{schertler2017towards}
		& 30.34 & 32.34 & 44.08 & 51.71 & 33.46 & 40.49 &   38.74    & 41.62 & 41.06 & 67.41 & 62.04 & 45.59 & 43.83 &   50.26  \\
		HSurf-Net~\cite{li2022hsurf}+ODP~\cite{metzer2021orienting}
		& 26.91 & 24.85 & 35.87 & 51.75 & 26.91 & 20.16 &   31.07    & 43.77 & 43.74 & 46.91 & 72.70 & 45.09 & 43.98 &   49.37  \\
		PCPNet~\cite{guerrero2018pcpnet}
		& 33.34 & 34.22 & 40.54 & 44.46 & 37.95 & 35.44 &   37.66    & 40.51 & 41.09 & 46.67 & 54.36 & 40.54 & 44.26 &   44.57  \\
		DPGO$^\ast$~\cite{wang2022deep}
		& 23.79 & 25.19 & 35.66 & 43.89 & 28.99 & 29.33 &   31.14    & - & - & - & - & - & - & -  \\
		NGLO~\cite{li2023NGLO}
		& 12.52 & 12.97 & 25.94 & \textbf{33.25} & 16.81 & 9.47  &   18.49    & 13.22 & 18.66 & \textbf{39.70} & 51.96 & 31.32 & 11.30 &   27.69  \\
		Ours
		& 10.28 & 13.23 & \textbf{25.40} & 35.51 & \textbf{16.40} & 17.92 &  19.79
		& 21.63 & 25.96 &         41.14  & 52.67 & \textbf{26.39} & 28.97 &  32.79  \\
		Ours-U+O
		& \textbf{10.26} & 13.47 & 25.85 & 36.04 & 16.54 & 17.95 &   20.02
		&         21.69  & 26.09 & 41.91 & 52.87 & 26.44 & 29.00 &   33.00 \\
		Ours-U+NGL
		&        {12.05} & \textbf{12.87} & 25.95 &        {33.43} & 16.44 & \textbf{8.97}  &   \textbf{18.29}
		& \textbf{12.30} & \textbf{18.20} & 39.96 & \textbf{51.57} & 31.11 & \textbf{10.56} &   \textbf{27.28} \\
		\bottomrule
	\end{tabular} }
    \vspace{-0.4cm}
\end{table*}
\begin{table}[t]
	\centering
	\footnotesize
	\setlength{\tabcolsep}{0.6mm}
	\caption{
		Unoriented normal RMSE on sparse point clouds with 3000 and 5000 points.
		Our approach has significant advantages.
	}
	\vspace{-0.4cm}
	\label{table:sparse}
	\resizebox{\linewidth}{!}{
		\begin{tabular}{c|ccccccccc}
			\toprule
			   & GraphFit & NeAF   & HSurf-Net  & NGLO   & Du \etal & CMG-Net & GCNO   & Ours   & Ours-U            \\  
			\midrule
			3K & 29.56    & 28.64  & 27.74      & 26.78  & 27.18    & 26.13   & 27.87  & 27.22  & \textbf{24.91}    \\  
			5K & 25.47    & 25.10  & 23.93      & 23.07  & 23.30    & 22.40   & 31.54  & 23.55  & \textbf{21.20}    \\  
			\bottomrule
		\end{tabular} }
	\vspace{-0.2cm}
\end{table}
\begin{table}[t]
    \footnotesize
    \centering
    \setlength{\tabcolsep}{0.6mm}
    \caption{
        Oriented normal RMSE on sparse point clouds.
        The algorithms of GCNO and PCA+MST run on the CPU, and the running time (seconds per 5000 points) of GCNO is much longer than other methods.
    }
    \vspace{-0.4cm}
    \resizebox{\linewidth}{!}{
    \begin{tabular}{c|cccccccc}
    \toprule
                  & \tabincell{c}{PCA\\+MST}
                  & \tabincell{c}{HSurf-Net\\+ODP}
                  & PCPNet
                  & GCNO
                  & NGLO
                  & Ours
                  & \tabincell{c}{Ours-U\\+O}
                  & \tabincell{c}{Ours-U\\+NGL}  \\
    \midrule
    3K     & 51.62                & 63.88        & 53.13     & 33.40      & 32.65       & 37.31    & 36.52       & \textbf{30.80}   \\
    5K     & 45.40                & 62.51        & 48.48     & 41.24      & 28.34       & 32.64    & 31.85       & \textbf{26.97}   \\
    Time   & \textbf{0.01+0.71}   & 3.87+31.75   & 3.34      & 822.60     & 0.10+2.82   & 3.54     & 1.44+3.54   & 1.44+0.10        \\
    \bottomrule
    \end{tabular} }
    \label{tab:sparse_o}
    \vspace{-0.2cm}
\end{table}
\begin{table}[t]
    \footnotesize
    \centering
    \setlength{\tabcolsep}{.6mm}
    \caption{
      Comparison of the \textit{unoriented} normal RMSE, the learnable network parameter (million) and the average inference time (seconds per 100K points) of different learning-based methods on the PCPNet dataset.
    }
    \vspace{-0.4cm}
    \resizebox{\linewidth}{!}{
    \begin{tabular}{l|cccccccccc}
      \toprule
                    & GraphFit & NeAF   & HSurf-Net  & NGLO        & Du \etal  & CMG-Net & Ours    & Ours-U         \\  
      \midrule
      RMSE          & 10.69    & 10.22  & 10.11      & 9.98        & 9.96      & 9.93    & 9.96    & \textbf{9.55}  \\  
      Param.        & 4.26     & 6.74   & 2.16       & 0.46+1.92   & 4.46      & 2.70    & 3.27    & \textbf{1.76}  \\  
      Time          & 292.12   & 400.81 & 72.47      & 0.56+70.77  & 295.69    & 109.98  & 65.89   & \textbf{9.77}  \\  
      \bottomrule
    \end{tabular}  }
    \label{tab:time_uo}
    \vspace{-0.2cm}
\end{table}
\begin{table}[t]
    \footnotesize
    \centering
    \setlength{\tabcolsep}{0.6mm}
    \caption{
        Comparison of the \textit{oriented} normal RMSE, the learnable network parameter (million) and the average inference time (seconds per 100K points) of different learning-based methods on the PCPNet dataset.
    }
    \vspace{-0.4cm}
    \resizebox{\linewidth}{!}{
    \begin{tabular}{l|ccccccc}
    \toprule
        & \tabincell{c}{HSurf-Net\\+ODP}
        & \tabincell{c}{AdaFit\\+ODP}
        & PCPNet
        & NGLO
        & Ours
        & \tabincell{c}{Ours-U\\+O}
        & \tabincell{c}{Ours-U\\+NGL}  \\
    \midrule
    RMSE          & 31.07         & 30.93         & 37.66  & 18.49         & 19.79   & 20.02       & \textbf{18.29}      \\
    Param.        & 2.16+0.43     & 4.87+0.43     & 22.36  & 0.46+1.92     & 3.27    & 1.76+3.27   & \textbf{1.76+0.46}  \\
    Time          & 72.47+236.35  & 56.23+248.54  & 63.02  & 0.56+70.77    & 65.89   & 9.77+65.89  & \textbf{9.77+0.56}  \\
    \bottomrule
    \end{tabular}  }
    \label{tab:time_o}
    \vspace{-0.2cm}
\end{table}

\begin{figure}[t]
	\centering
	\includegraphics[width=.95\linewidth]{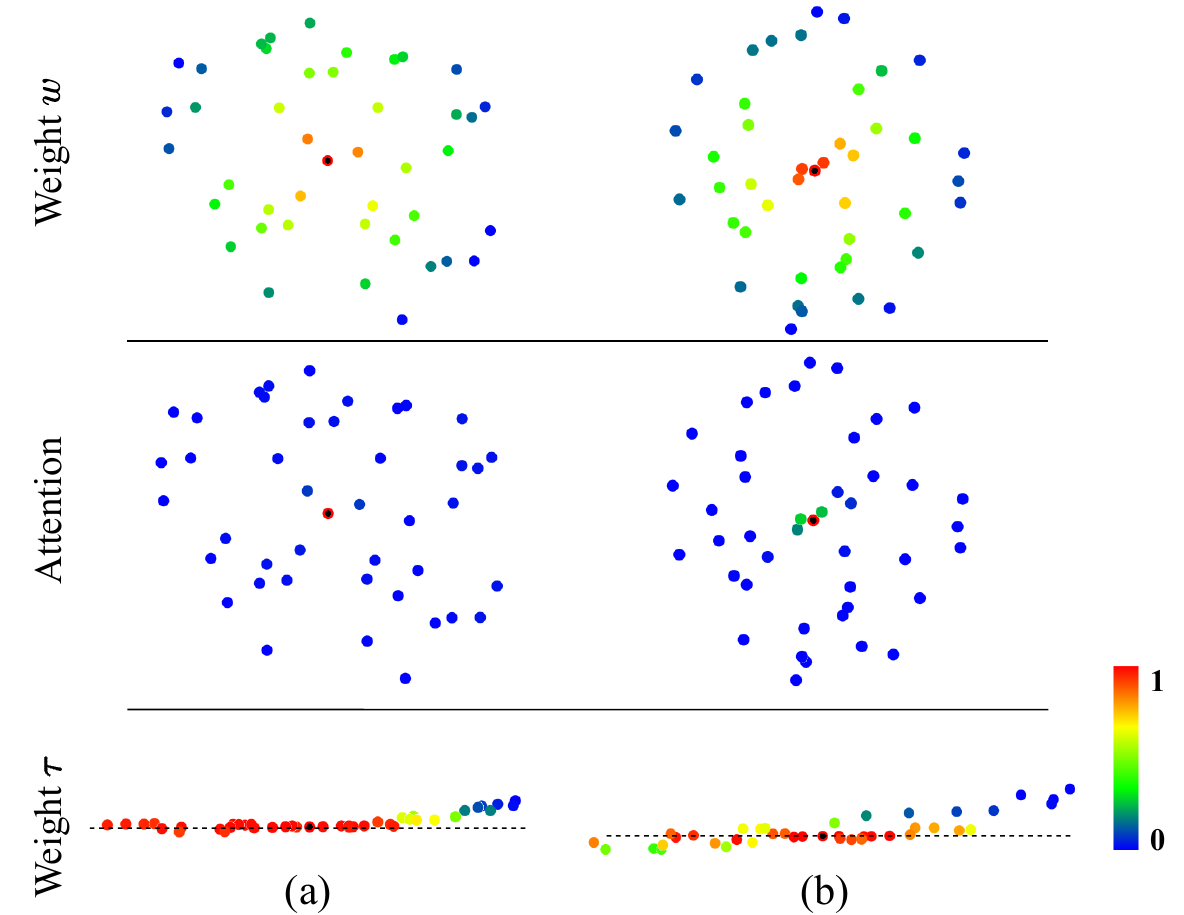}  \vspace{-0.3cm}
  \caption{
    Visualization of the learned weight $w$, attention weight, and weight $\tau$ in two point cloud patches (a)(b).
    They show the points that the model focuses on at different stages of the normal estimation process.
    Specifically, these points are the following three types: (top) the points closer to the center, (middle) the query point and its neighbors, and (bottom) the points coplanar with the query point.
    The red color indicates that the point has a large value, while the blue color indicates that it has a small value.
    The black point is the query point of the patch.
    The first two rows are top views of the patch, and the third row is a side view.
    The viewing angle of the third row is changed for better visualization.
  }
  \label{fig:weight_attn}
\end{figure}

\begin{figure}[t]
  \centering
  \includegraphics[width=\linewidth]{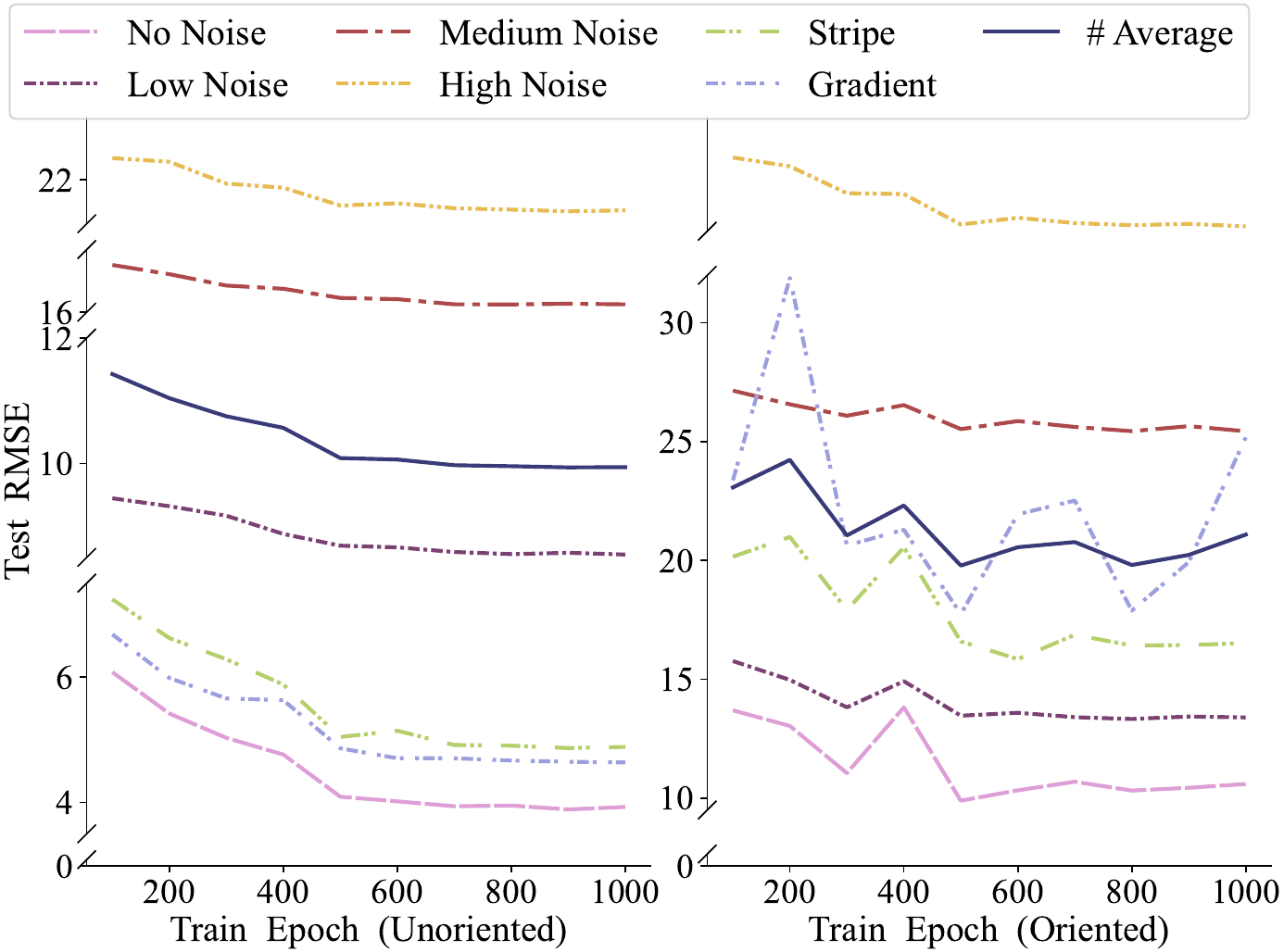}  \vspace{-0.6cm}
  \caption{
    Unoriented and oriented normal evaluation results on the PCPNet test set using our models trained for 100 to 1000 epochs.
    We report the RMSE results under different noise levels and density variations along with their average values.
    Note that the normals used in the unoriented and oriented normal evaluation are the same, but evaluated using different metrics, \ie, ${\rm RMSE}_{\rm unoriented}$ and ${\rm RMSE}_{\rm oriented}$.
  }
  \label{fig:test_epoch}
  \vspace{-0.2cm}
\end{figure}

\begin{table*}[t]
\centering
\footnotesize
\setlength{\tabcolsep}{1.5mm}
\caption{
    Ablation studies for oriented normals on the PCPNet dataset.
    The average results under unoriented metric are also provided in the last column.
}
\vspace{-0.4cm}
\label{tab:ablation}
\resizebox{\linewidth}{!}{
\begin{tabular}{ll|cccc||cccc|cc| >{\columncolor{mygray}} c || >{\columncolor{mygray}} c}
    \toprule
    \multicolumn{2}{c|}{\multirow{2}{*}{\textbf{Ablation}}} & \multirow{2}{*}{\shortstack{Feat.\\Enco.}} & \multirow{2}{*}{\shortstack{Module \\ $\mathcal{H}$}}
                                                            & \multirow{2}{*}{\shortstack{Loss}} & \multirow{2}{*}{\shortstack{Point \\ Samp.}}
                                                            & \multicolumn{4}{c|}{Noise} & \multicolumn{2}{c|}{{Density}} &   &  \\
    & & & & & & None & 0.12\% & 0.6\% & 1.2\% & Stripe & Gradient & \multirow{-2}{*}{\shortstack{Oriented \\ Average}} & \multirow{-2}{*}{\shortstack{Unoriented \\ Average}} \\
    \midrule
    \multirow{3}{*}{\textbf{(a)}}
    & w/o patch encoding  & & \checkmark & \checkmark & \checkmark & 35.19 & 42.23 & 55.59 & 61.38 & 38.92 & 41.49 &   45.80 & 18.83 \\
    & w/o shape encoding  & & \checkmark & \checkmark & \checkmark & 69.72 & 64.37 & 81.87 & 77.07 & 74.84 & 90.35 &   76.37 & 14.94 \\
    & w/o weight $w$      & & \checkmark & \checkmark & \checkmark & 11.15 & 14.32 & 26.49 & 36.03 & 17.99 & 26.03 &   22.00 & 10.48 \\
    \hline
    \multirow{1}{*}{\textbf{(b)}}
    & w/o module $\mathcal{H}$ & \checkmark & & \checkmark & \checkmark & 12.08 & 14.53 & 25.87 & 35.88 & 18.45 & 31.84 &   23.11 & 10.24 \\
    \hline
    \multirow{2}{*}{\textbf{(c)}}
    & w/o $\mathcal{L}_{sin}, \mathcal{L}_{sgn}$ & \checkmark & \checkmark & & \checkmark & 23.86 & 25.55 & 34.13 & 42.48 & 32.42 & 41.30 &   33.29 & 20.23 \\
    & w/o $\mathcal{L}_{mse}$                    & \checkmark & \checkmark & & \checkmark & 18.89 & 29.83 & 35.61 & 43.53 & 24.41 & 33.71 &   30.99 & 10.03 \\
    \hline
    \multirow{6}{*}{\textbf{(d)}}
    & w/o density gradient          & \checkmark & \checkmark & \checkmark & & 12.10 & 18.25 & 28.05 & 38.15 & 19.79 & 28.09 &   24.07 & 10.00 \\
    & w/o random sample             & \checkmark & \checkmark & \checkmark & & 11.01 & 13.79 & 25.64 & 35.86 & 17.22 & 25.71 &   21.54 & 9.94  \\
    & $\zeta\!=\!1/2$               & \checkmark & \checkmark & \checkmark & & 10.99 & 14.04 & 25.66 & 35.78 & 17.73 & 37.82 &   23.67 & \textbf{9.92} \\
    & $\zeta\!=\!1/3$               & \checkmark & \checkmark & \checkmark & & 13.27 & 15.42 & 26.82 & 37.16 & 17.52 & 28.11 &   23.05 & 9.95 \\
    & $N_{\boldsymbol P}\!=\!1100$  & \checkmark & \checkmark & \checkmark & & 10.67 & 14.21 & 25.54 & 35.97 & 16.80 & 26.98 &   21.69 & 9.99 \\
    & $N_{\boldsymbol P}\!=\!1300$  & \checkmark & \checkmark & \checkmark & & 12.44 & 14.53 & 25.93 & 35.79 & 18.40 & 19.85 &   21.16 & 9.98 \\
    \hline
    & \textbf{Final} & \checkmark & \checkmark & \checkmark & \checkmark &
    \textbf{10.28} & \textbf{13.23} & \textbf{25.40} & \textbf{35.51} & \textbf{16.40} & \textbf{17.92} &   \textbf{19.79}  & {9.96} \\
    \bottomrule
\end{tabular} }
\vspace{-0.3cm}
\end{table*}

\subsection{Complexity and Efficiency}

In this evaluation experiment, we compare the learning-based methods on the same machine with an NVIDIA 2080 Ti GPU.
The comparative experiment is first conducted on the task of unoriented normal estimation, and the learning-based methods that have good performance on the PCPNet dataset are selected as baselines, such as GraphFit~\cite{li2022graphfit}, NeAF~\cite{li2023NeAF} and Du \etal~\cite{du2023rethinking}.
As shown in Table~\ref{tab:time_uo}, we report the average RMSE of unoriented normal on the PCPNet dataset, the number of learnable network parameters, and the execution time for unoriented normal estimation.
Our method achieves significant performance improvement with minimal parameters and running time.
It is worth noting that our running efficiency is dozens or even hundreds of times faster than other baseline methods, such as $9.77$ (Ours-U) \vs $~295.69$ (Du \etal~\cite{du2023rethinking}).

In the oriented normal evaluation experiments, we compare our method to PCPNet~\cite{guerrero2018pcpnet}, DPGO~\cite{wang2022deep}, NGLO~\cite{li2023NGLO} and other methods that are based on a two-stage paradigm.
We make different combinations of existing works to estimate oriented normals, such as AdaFit+ODP and HSurf-Net+ODP.
Among the baseline methods, AdaFit~\cite{zhu2021adafit}, HSurf-Net~\cite{li2022hsurf} and ODP~\cite{metzer2021orienting} are learning-based methods, and the others are traditional methods.
In Table~\ref{tab:time_o}, we report the average RMSE of oriented normal on the PCPNet dataset, the number of learnable network parameters, and the execution time of each method for oriented normal estimation.
Our method achieves a large performance improvement with relatively fewer parameters and less running time.

\subsection{Visualization of Weight and Attention}
As shown in Fig.~\ref{fig:weight_attn}, we visualize the learned attention weight in the normal prediction module $\mathcal{H}$, weight $\tau$ in Eq.~\eqref{eq:output} and weight $w$ in Eq.~\eqref{eq:weight}.
They illustrate the points that the model focuses on at different stages of the normal estimation process.
The weight $w$ indicates that the model focuses on points closer to the center during the patch and shape encoding.
The weight $\tau$ indicates that the model focuses on points coplanar with the query point during the final local feature modulation for normal prediction.
The attention weight indicates that the model focuses on the query point during the final oriented normal prediction of the query point.

\subsection{Ablation Studies}  \label{sec:ablation}

We provide ablation results for oriented normal estimation in Table~\ref{tab:ablation} (a)-(d), which are discussed as follows.

\noindent\textbf{(a) Feature Encoding}.
(i) We realize the oriented normal estimation without using the patch encoding or the shape encoding.
(ii) The distance-based weight $w$ is not used in both patch encoding and shape encoding.

\noindent\textbf{(b) Module $\mathcal{H}$}.
The attention-weighted normal prediction module $\mathcal{H}$ is replaced with simple MLP layers.

\noindent\textbf{(c) Losses $\mathcal{L}_{sin}, \mathcal{L}_{sgn}$ and $\mathcal{L}_{mse}$}.
In our pipeline, we regress the unoriented normal and its orientation sign of the query point $q$, and constrain them in loss functions $\mathcal{L}_{sin}$ and $\mathcal{L}_{sgn}$, respectively.
Here, we do not use $\mathcal{L}_{sin}$ and $\mathcal{L}_{sgn}$, and directly predict the oriented normal $\vec{\mathbf{n}}_q$ of the point $q$ from surface embedding and compute its MSE loss.
Moreover, we also conduct an experiment by removing the weighted mean square error loss of neighboring point normals in Eq.~\eqref{eq:loss_n} to show its effect on the estimation of query point normal, \ie, without using $\mathcal{L}_{mse}$.

\noindent\textbf{(d) Point Sampling}.
In the shape encoding, we obtain a global point set ${\boldsymbol P}_q$ by a probability-based sampling strategy as in Eq.~\eqref{eq:sample}, which includes density gradient term and random sample term.
The point set ${\boldsymbol P}_q$ includes $N_{\boldsymbol P} \!=\! 1200$ points, and the ratio of randomly sampled points is $\zeta \!=\! 1/1.5$.
(i) We only adopt one of the two terms in Eq.~\eqref{eq:sample} for point sampling, \eg, "w/o density gradient" means all points are randomly sampled and "w/o random sample" means all points are sampled by the density gradient.
(ii) The ratio $\zeta$ is changed to $1/2$ and $1/3$.
(iii) The number of points $N_{\boldsymbol P}$ is set to $1100$ and $1300$.

From Table~\ref{tab:ablation}, we can conclude that both patch encoding and shape encoding are vital for learning accurate oriented and unoriented normals in our pipeline.
The adoption of the weight $w$ and the attention-weighted normal prediction module $\mathcal{H}$ effectively improves the algorithm's performance.
Compared to directly predicting the oriented normal $\vec{\mathbf{n}}_q$, solving it separately (normal and its sign) by learning signed hyper surfaces is significantly better.
The constraints on neighboring point normal help the network model fully explore the local and global geometry of point cloud patches, and ensure the local normal consistency, especially in oriented normal estimation.
The combination of the density gradient term and the random sample term in sampling can produce better results than either one alone.
The proportion and number of points in the sampling of shape encoding have a significant positive effect on oriented normals, but very little on unoriented normals.

\noindent\textbf{(e) Epoch of Model Training}.
To determine how many epochs a model needs to train, we evaluate our method on the PCPNet test set using models trained for 100 to 1000 epochs.
The estimated normals are measured using the evaluation metrics ${\rm RMSE}_{\rm unoriented}$ and ${\rm RMSE}_{\rm oriented}$ of normal angles.
The evaluation results are shown in Fig.~\ref{fig:test_epoch}.
We provide the results at different noise levels and different density variations along with their average results.
It can be seen from the curves in the figure that the errors of unoriented normal evaluation keep decreasing, while the errors of oriented normal evaluation fluctuate greatly.
In our training, we observed that the model is harder to converge in oriented normal estimation than in unoriented normal estimation.
After about 800 epochs of training, the errors of oriented normal evaluation reach a minimum value, and the errors of unoriented normal evaluation also reach the lowest value and remain unchanged.
Therefore, in all experiments of the paper, we use the model trained in 800 epochs.

\begin{figure}[t]
  \centering
  \includegraphics[width=\linewidth]{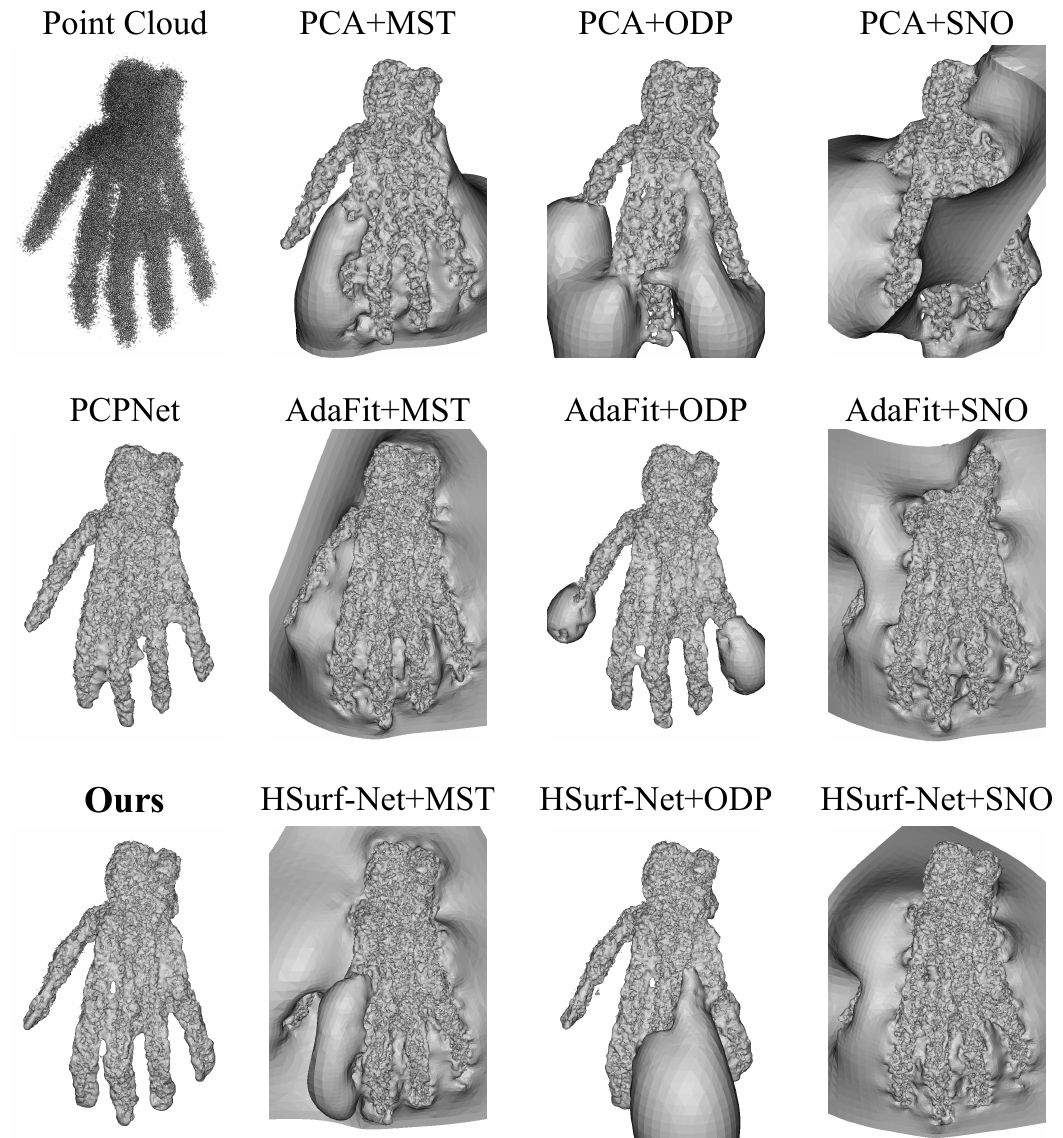}  \vspace{-0.8cm}
  \caption{
    Comparison of surface reconstruction results from a noisy point cloud using oriented normals estimated by different methods.
  }
  \label{fig:poissonRecon_hand}
  \vspace{-0.2cm}
\end{figure}

\begin{figure*}[t]
  \centering
  \subfigure[Street scene 1 of the KITTI dataset.]{
    \includegraphics[width=\linewidth]{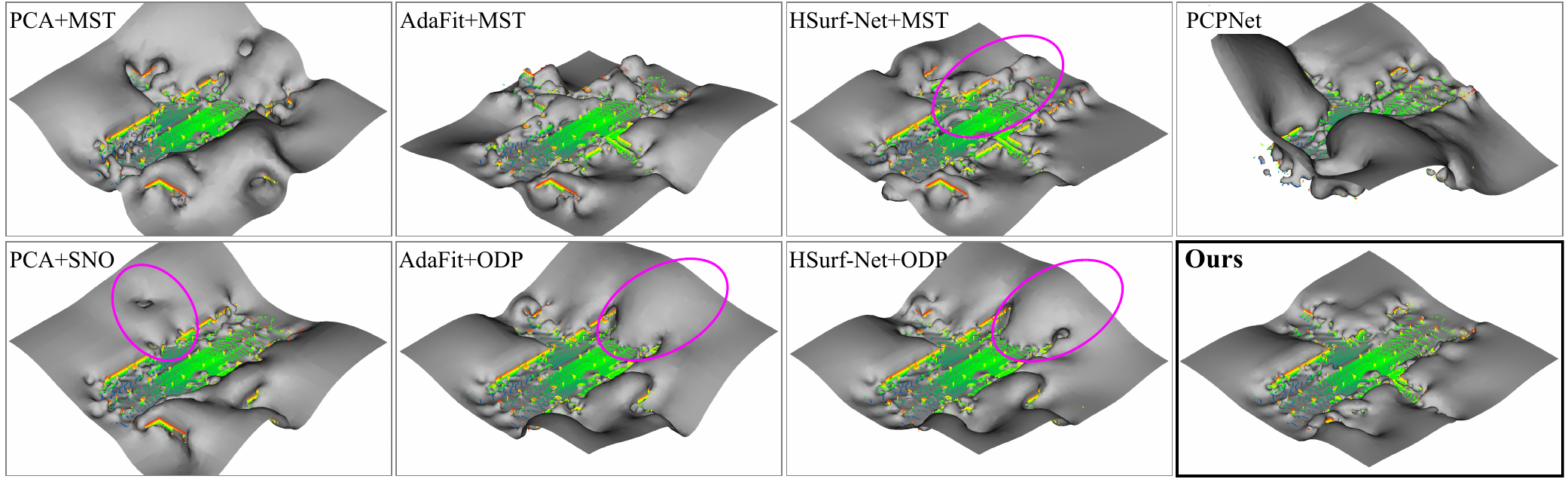}  \vspace{-0.5cm}
    \label{fig:KITTI_06_600}
  }
  \subfigure[Street scene 2 of the KITTI dataset.]{
    \includegraphics[width=\linewidth]{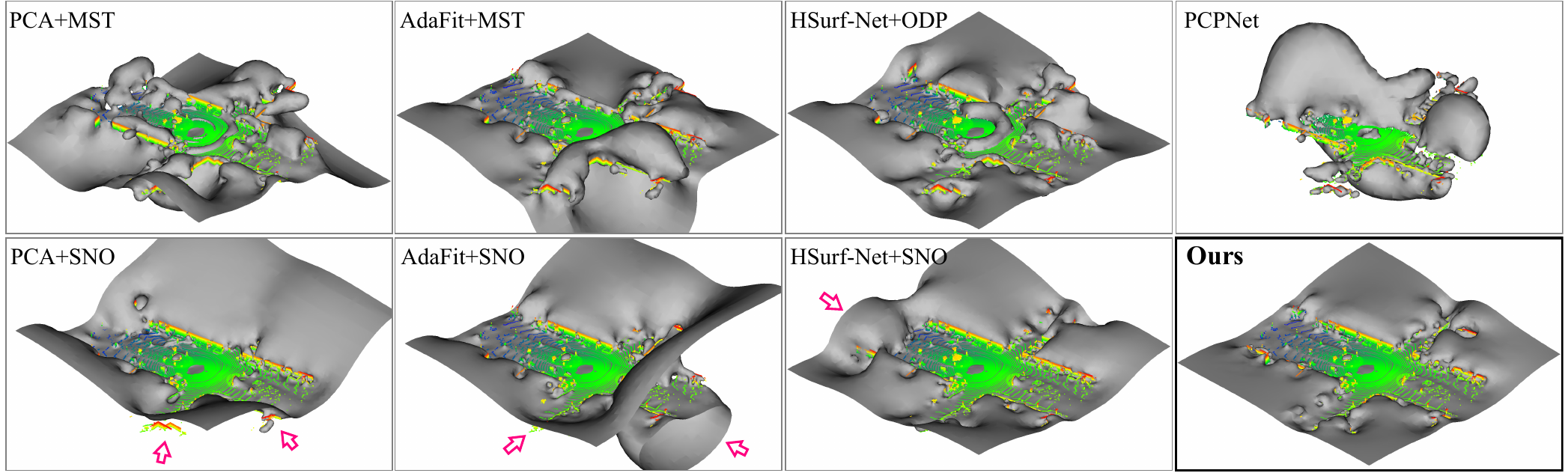}  \vspace{-0.5cm}
    \label{fig:KITTI_06_700}
  }
  \vspace{-0.4cm}
  \caption{
    Comparison of reconstructed surfaces using oriented normals estimated by different methods on the sparse and non-uniformly distributed point clouds of the KITTI dataset.
    The raw point cloud has an open surface structure and is colored with height values.
    The reconstructed surface of the street scene is shown in gray.
  }
  \label{fig:KITTI_recons}
  \vspace{-0.2cm}
\end{figure*}

\begin{figure*}[t]
	\centering
	\includegraphics[width=\linewidth]{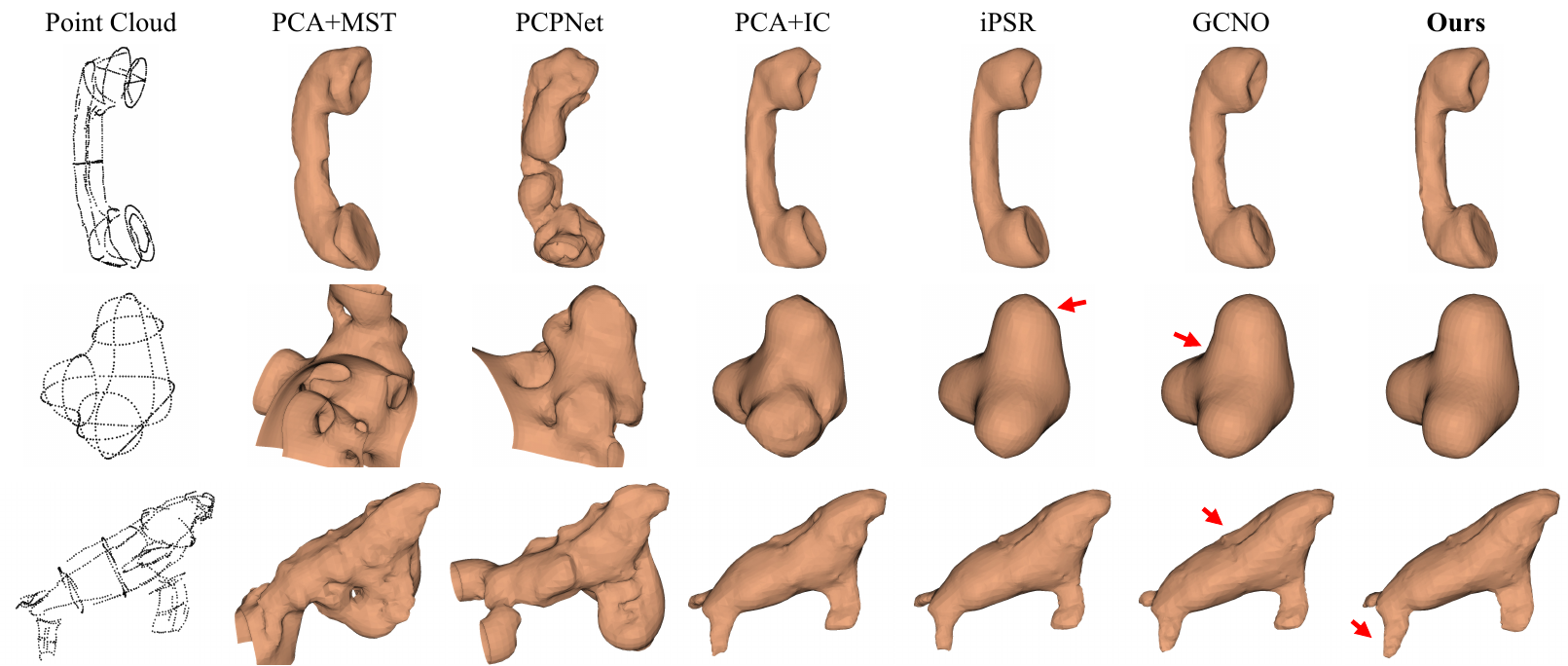}  \vspace{-0.8cm}
  \caption{
    Oriented normal estimation on wireframe point clouds with sparse and non-uniform sampling.
    We compare shape surfaces generated by Poisson surface reconstruction.
    The number of points in the input point cloud is less than 1000.
    The ground truth is not available.
  }
  \label{fig:wireframe}
  \vspace{-0.2cm}
\end{figure*}

\begin{figure*}[t]
  \centering
  \includegraphics[width=\linewidth]{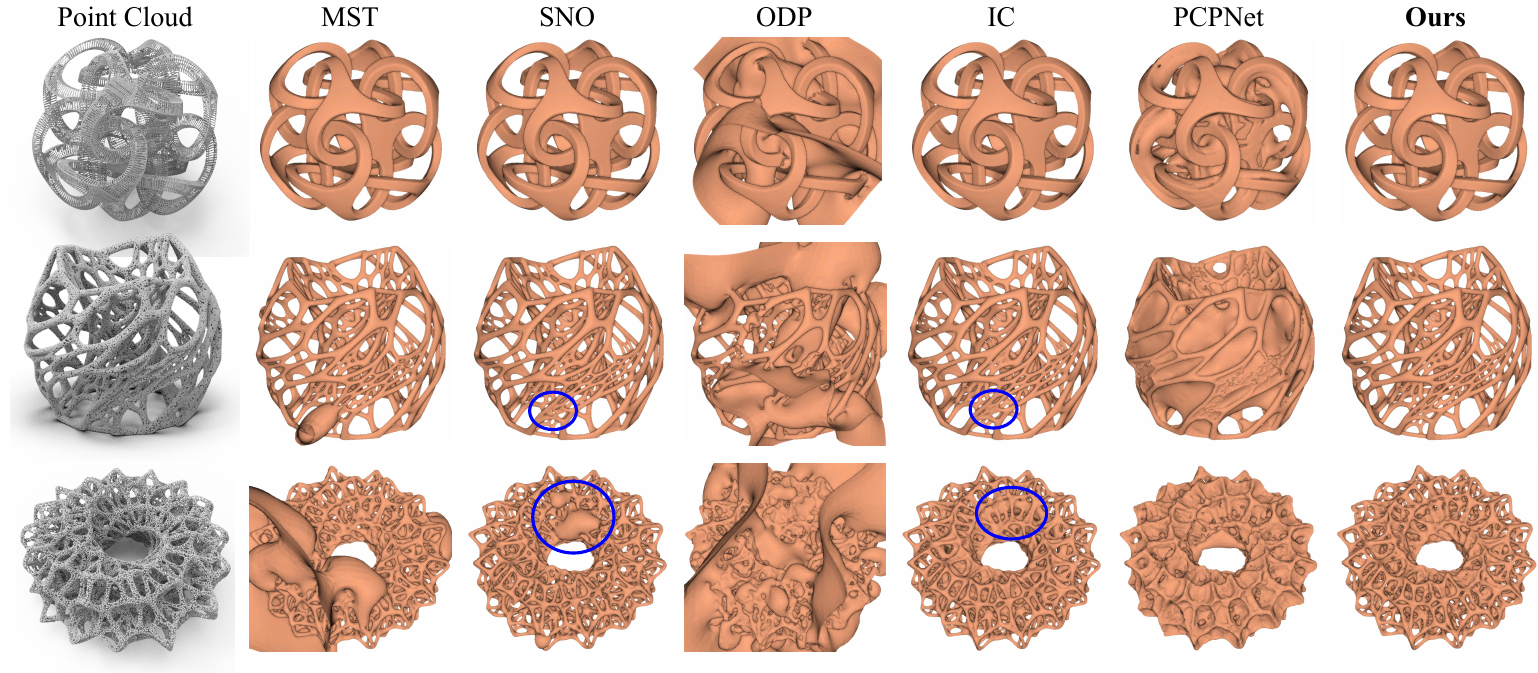}  \vspace{-0.95cm}
  \caption{
    Oriented normal estimation on point clouds with highly complex topology/geometry.
    The initial unoriented normals for methods MST~\cite{hoppe1992surface}, SNO~\cite{schertler2017towards}, ODP~\cite{metzer2021orienting} and IC~\cite{xiao2023point} are provided by PCA.
    The point cloud in the first row has about 70K points, while the point clouds in the second and third rows have 100K points.
    The surfaces are reconstructed using the estimated oriented normals.
    Our method can provide explicit shape structures from these three point clouds and achieves good results on the first two point clouds.
    However, all methods fail to distinguish the details inside the model in the third point cloud.
  }
  \label{fig:poissonRecon_NestPC}
  \vspace{-0.2cm}
\end{figure*}

\begin{figure*}[t]
	\centering
	\includegraphics[width=\linewidth]{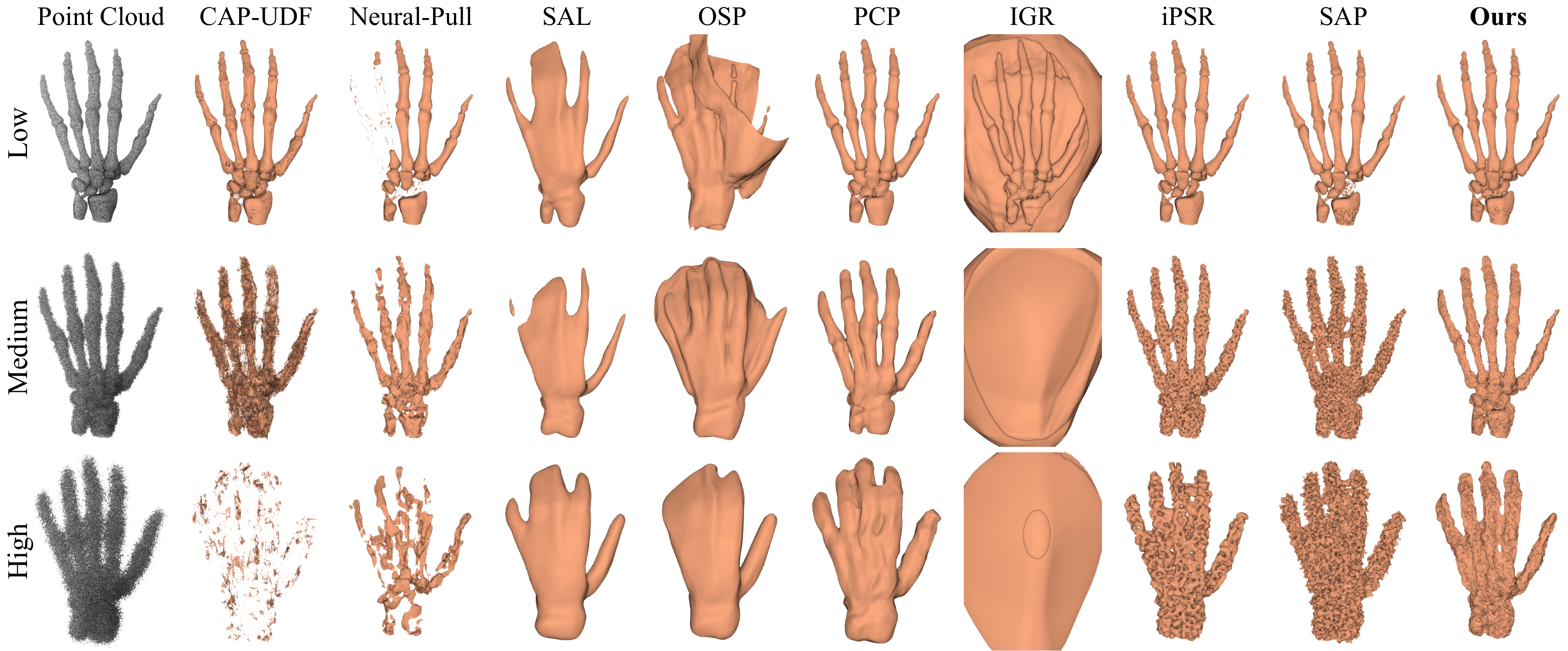}  \vspace{-0.8cm}
  \caption{
    Comparison with surface reconstruction methods.
    The input point clouds are with different noise levels (low, medium and high).
    As the noise increases, our method has the advantage of better performance.
  }
  \label{fig:compRecon_hand}
\end{figure*}

\begin{figure*}[t]
  \centering
  \includegraphics[width=\linewidth]{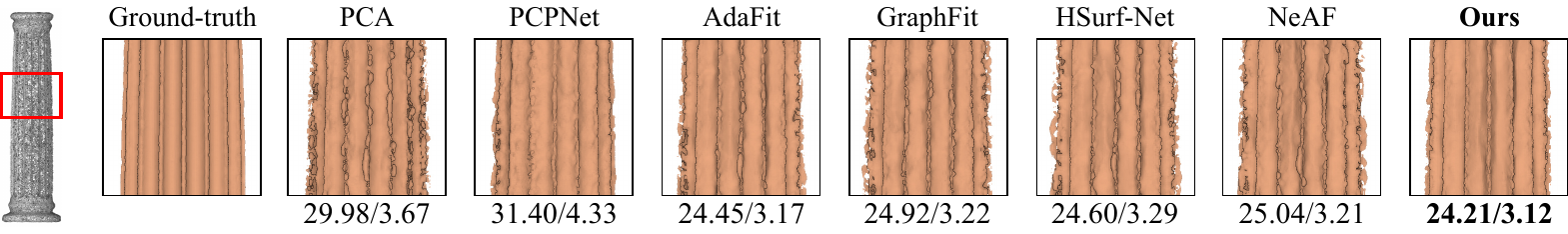}  \vspace{-0.65cm}
  \caption{
    Comparison of point cloud filtering using normals estimated by different methods.
    The first column shows the raw point cloud (grey), the surface reconstructed using its filtered point cloud is shown behind with a local zoomed-in view.
    For a quantitative comparison, the average RMSE of the estimated normals for each point cloud and the Chamfer Distance (CD, $\times10^{-5}$) of the filtered point cloud are reported separately below the shape in RMSE/CD format.
  }
  \label{fig:filtering_l}
\end{figure*}

\section{Applications}
In the following experiments, we will demonstrate that our method can accurately estimate normals on point clouds with noise, density variations, and complex geometries, thereby facilitating downstream tasks, such as surface reconstruction and point cloud filtering.

\subsection{Surface Reconstruction}

\subsubsection{Oriented Normal Estimation Methods}
For surface reconstruction, we first compare our method with oriented normal estimation methods.
Based on the oriented normals estimated by different methods, we use the Poisson reconstruction algorithm~\cite{kazhdan2013screened} to reconstruct surfaces from point clouds.
In Fig.~\ref{fig:poissonRecon_hand}, we show a visual comparison of the reconstructed surfaces on a noisy point cloud.
We can see that our method helps the Poisson algorithm to reconstruct better surfaces from point clouds with noise and complex geometries compared to the baseline methods.

\noindent\textbf{Real-world LiDAR Data}.
To verify the generalization ability of our method on LiDAR point clouds of outdoor scenes, we test directly on the KITTI dataset~\cite{geiger2012we} with the network model trained on the PCPNet dataset.
The point clouds in this dataset have non-uniform density and open surface structure, which pose a great challenge for oriented normal estimation.
We only report qualitative results on this dataset as it does not provide the ground truth normals or surfaces.
As shown in Fig.~\ref{fig:KITTI_recons}, we use the Poisson surface reconstruction algorithm~\cite{kazhdan2013screened} to generate surfaces using oriented normals estimated by different methods.
As can be seen in the figures, compared to the baselines, our estimated normals facilitate the algorithm to reconstruct surfaces that can more accurately depict the spatial structure and distribution of real scenes.

\noindent\textbf{Wireframe Point Clouds}.
The point cloud used above can accurately describe the details of objects or scenes.
In addition, we can also use wireframe point clouds to describe the outline of objects, which provide a compact skeletal representation with a very small number of points.
Due to the extremely sparse and non-uniform distribution of the data, restoring 3D surfaces from such point cloud data has always been very challenging.
As shown in Fig.~\ref{fig:wireframe}, we visualize the reconstructed surfaces of the baseline methods, such as PCA+MST~\cite{hoppe1992surface}, PCPNet~\cite{guerrero2018pcpnet} and GCNO~\cite{xu2023globally}.
They represent three typical classes of oriented normal estimation methods, namely two-stage pipeline, learning-based single-stage, and traditional scheme-based single-stage.
Both Isovalue Constraint (IC)~\cite{xiao2023point} and iPSR~\cite{hou2022iterative} can serve as a kind of improved Poisson surface reconstruction, and their optimization of implicit surfaces can also solve oriented normals with globally consistent orientation.
The initial unoriented normals of IC~\cite{xiao2023point} are estimated by PCA, \ie, PCA+IC in Fig.~\ref{fig:wireframe}, and iPSR~\cite{hou2022iterative} uses randomly initialized point normals.
The experimental results show that our method is capable of handling the wireframe-type inputs, and is superior to some latest competitors in certain details.

\noindent\textbf{Complex Topological Data}.
As shown in Fig.~\ref{fig:poissonRecon_NestPC}, we provide several point clouds with nested structures whose highly complex topology/geometry poses great challenges for normal orientation.
Our method shows significant performance improvement over some baseline methods, especially the deep learning-based counterpart, PCPNet.
The propagation-based methods do not use globally sampled points to determine the orientation, but instead rely on the orientation propagation of neighboring point normals.
Their scheme does not have advantages in some regions with adjacent surfaces or large curvature changes.

\subsubsection{Surface Reconstruction Methods}
To further evaluate the effect of estimated normals in surface reconstruction, we compare our method with other methods that are designed for surface reconstruction from point clouds.
These methods determine the zero-level set of the learned implicit function via the signed distance field, and use the marching cubes algorithm~\cite{lorensen1987marching} to extract a surface of the point cloud.
The baseline methods include Neural-Pull~\cite{ma2020neural}, CAP-UDF~\cite{Zhou2022CAP-UDF}~\cite{zhou2024cap}, OSP~\cite{ma2022reconstructing}, PCP~\cite{ma2022surface}, SAL~\cite{atzmon2020sal}, IGR~\cite{gropp2020implicit} and Shape As Points (SAP)~\cite{peng2021shape} and their distance fields are predicted through a learning-based pipeline.
A visual comparison of the extracted surfaces on point clouds with different noise levels is shown in Fig.~\ref{fig:compRecon_hand}.
We can see that, based on the accurate normals estimated by our method, the Poisson reconstruction algorithm~\cite{kazhdan2013screened} generates more complete and detailed geometry from noisy point clouds than baseline methods.

\subsection{Point Cloud Filtering}
Point cloud data collected from the real world are often noisy due to sensors and environments.
Therefore, point cloud filtering is often an important preprocessing step before further processing of point clouds.
In this evaluation, we use the algorithm proposed in \cite{lu2020low} to perform point cloud filtering using the estimated point normals.
As shown in Fig.~\ref{fig:filtering_l}, we provide the qualitative and qualitative comparison results of the filtered point clouds and their surfaces reconstructed by Poisson surface reconstruction algorithm~\cite{kazhdan2013screened}.
It can be seen that the filtering algorithm can benefit from our estimated normals, and it smooths the surfaces in flat areas while still keeping detailed structures at sharp edges.

\section{Conclusion}

In this work, we formulate the oriented normal estimation of point clouds as the learning of signed hyper surfaces.
We first review the explicit surface fitting and the implicit surface learning, and derive the formulation of the signed hyper surfaces from their inspiration.
Then, we propose to use an attention-weighted normal prediction module to recover the normal and its sign of the query point from the embedding of the signed hyper surfaces.
Finally, we introduce how such surfaces can be learned from the patch encoding and shape encoding using the designed loss functions.
We conduct extensive evaluation and ablation experiments to report the state-of-the-art performance and justify the effectiveness of our designs.
We show that oriented normal estimation is tightly coupled with surface reconstruction, and that our estimated normals can facilitate surface reconstruction algorithms to generate better object structures.

For the normal estimation task, we make a comprehensive analysis of the problem in theory and make specific designs in technology.
It is these important innovative designs that enable our method to outperform the state-of-the-art methods in both unoriented and oriented normal estimation on the widely used benchmarks.
In summary, we explore a new idea for learning local features and geometric properties from point clouds.
It can better serve the community for point cloud processing and has a positive impact on the performance improvement of downstream tasks using normals.
Future work includes developing noise-adaptive techniques to handle more diverse point clouds and integrating our method into recently developed surface reconstruction methods.
In addition, the transfer of contextual information between adjacent points or patches is a research direction worth exploring.
The limitations of our approach are discussed in the supplementary material, and some failure cases are also provided.


\bibliographystyle{IEEEtran}
\bibliography{egbib}



\end{document}